\documentclass[11pt]{article}
\usepackage[letterpaper,margin=1in]{geometry}
\usepackage{times,natbib}
\usepackage[T1]{fontenc}
\usepackage[utf8]{inputenc}
\usepackage{amsmath,amssymb,amsthm}
\newtheorem{proposition}{Proposition}
\newtheorem{theorem}{Theorem}
\usepackage{graphicx,booktabs,array,calc,longtable}
\usepackage{microtype}
\usepackage{hyperref,url}

\hypersetup{hidelinks,pdfauthor={Manli Yan, Yaowen Yu, Yong Zhao, Yuebin Lin, Shoudong Han},pdftitle={Do Neural Networks Preserve Case Structure? Case-Based Decomposition, Interpretation, and Decision Consistency}}
\title{Do Neural Networks Preserve Case\\ Structure? Case-Based Decomposition,\\ Interpretation, and Decision Consistency}
\author{Manli Yan \quad Yaowen Yu \quad Yong Zhao\\
Yuebin Lin \quad Shoudong Han\\[0.5em]
\small School of Artificial Intelligence and Automation,\\
\small Huazhong University of Science and Technology}
\date{}
\begin{document}
\maketitle

\begin{abstract}
Neural networks increasingly inform consequential decisions, making their reliability increasingly important. Yet their internal mechanisms provide little evidence of whether decisions remain grounded in the training cases and which cases ultimately support or oppose their outcomes. Without this connection between decisions and training cases, users cannot determine whether a model has learned reliable decision patterns from data. This motivates a fundamental question: do neural networks preserve case structure? We establish a connection between neural networks and Case-Based Decision Theory (CBDT), showing that trained neural networks can preserve a recoverable case structure through their learned representations. Such a structure allows fitted decision margins to be decomposed into individual case contributions. We identify the conditions under which this recovered case structure admits a CBDT interpretation. We further establish decision consistency between this interpretation and the decision selected by the original neural network. Experiments on a controlled CBDT setting and three decision tasks based on real-world data validate our approach. These results connect neural network decisions with the cases that shape them. This connection allows model choices to be traced back to supporting and opposing cases, providing a basis for assessing the reliability of neural network decisions.
\end{abstract}

\section{Introduction}\label{introduction}

Neural networks learn prediction and decision functions from training cases through internal representations \citep{Bengio2013,LeCun2015,Goodfellow2016}. Prior work studies what these representations encode and how they vary across trained networks and learning \citep{Li2016,Guo2024,Chen2024}. However, these representations remain hidden during inference, which typically exposes only scores or a selected action, not which training cases support or oppose that choice or by how much. This missing connection between decisions and training cases makes the decision harder to assess and raises concerns about the reliability of these decisions.

Existing work has studied how neural-network decisions are influenced by input factors, learned representations, and training cases. At the input level, feature-attribution methods identified influential features through perturbations or additive attributions \citep{Ribeiro2016,Lundberg2017,Sundararajan2017}. Sanity checks further examined whether these explanations depended on the learned model and data \citep{Adebayo2018}. At the representation level, concept-based methods identified human-interpretable concepts associated with a decision \citep{Koh2020,Xu2024,Ismail2024}, while sparse-feature and circuit analyses sought latent features and causal mechanisms learned by the network itself \citep{Huben2024,Marks2025}. At the training-case level, Influence Functions and TracIn estimated how individual training examples affected model outputs through local perturbations or optimization trajectories \citep{Koh2017,Pruthi2020}. Related approaches studied how outputs varied across alternative training sets or traced example effects through approximate optimization unrolling \citep{Ilyas2022,Bae2024}. Recent studies further extended data attribution to parameter-efficient language models, diffusion models, and larger computational settings \citep{Kwon2024,Zheng2024,Schioppa2024}. They also showed that attribution patterns changed with training stage, model scale, traced behavior, access to the training process, and parameter weighting \citep{Wang2025,Khaddaj2025,Chang2025,Wei2025,Lee2026,Li2026}. Together, these methods have made training cases visible in model analysis, while how such cases jointly form neural-network decisions has not been systematically characterized.

A case-based view offers a natural way to formulate this missing connection. CBDT evaluates each candidate action using the action-specific returns associated with relevant past cases, with more relevant cases receiving greater weight \citep{Gilboa1995,Gilboa2012}. Case-Based Reasoning (CBR) likewise uses previous cases to reason about new problems \citep{Kolodner1993,Aamodt1994}. In neural networks, training-point decompositions, prototype models, and surrogate methods also made individual examples visible in prediction or explanation \citep{Yeh2018,Chen2019,Xue2024,Engel2024}. Together, these lines have shown that cases can ground decisions and neural predictions can be related to examples, leaving open whether trained neural networks preserve structures that connect their decisions to the cases underlying those decisions. Here, case structure denotes the organization of case-level contributions linking a model decision to the training cases involved in learning.

\begin{figure}[tb]
\centering
\begin{minipage}[b]{1.0\linewidth}\centering
\includegraphics[width=0.95\linewidth]{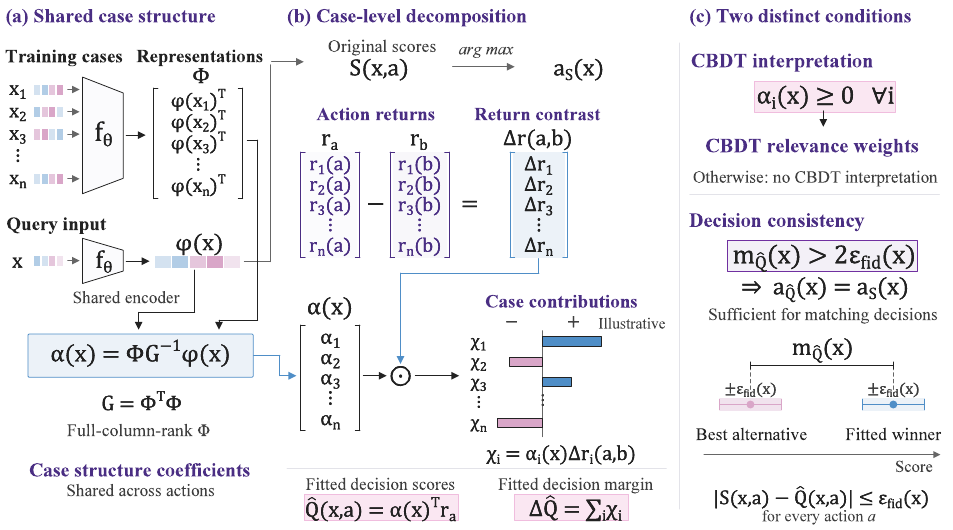}
\end{minipage}
\caption{Overview of the case structure underlying neural network decisions. The learned representation connects neural network decisions to case-level contributions, enabling CBDT interpretation and consistency verification with the original model decisions.}\label{fig:1}
\end{figure}

In this work, we study whether neural networks preserve such a case structure and establish its decomposition, interpretation, and decision consistency. Our contributions are threefold (Figure~\ref{fig:1}):

\begin{itemize}
\item
  First, we show that trained neural networks can preserve a recoverable case structure through their learned representations. This structure enables the decomposition of neural network decisions into individual case contributions.
\item
  Second, we establish when the recovered case structure obtains a CBDT interpretation. This condition distinguishes valid case decomposition from case-based interpretation of neural network decisions.
\item
  Third, we establish decision consistency between this interpretation and the decision selected by the original neural network. This connection links case-level contributions to the final model choice and makes supporting and opposing cases traceable.
\end{itemize}

\section{Background and Problem Setup}\label{background-and-problem-setup}

Neural network decisions and case-based decisions provide complementary views of how decisions can be understood. This section introduces the theoretical background and problem setup used throughout the paper. We first introduce the case-based view of decision making in Section 2.1, then describe neural network decision models in Section 2.2, and finally formulate the connection between neural network decisions and case-level contributions in Section 2.3.

\subsection{Case-Based Decision Theory}\label{case-based-decision-theory}

CBDT models decisions through previous cases and their associated outcomes \citep{Gilboa1995}. It evaluates how prior cases relate to the current situation and shape preferences among alternative choices.

Let $\mathcal{D}=\{(x_i,\mathbf r_i)\}_{i=1}^{n}$ denote $n$ cases, where $x_i$ represents the observed information of case $i$, and $\mathbf{r}_i=(r_i(a))_{a\in\mathcal{A}}\in\mathbb{R}^{|\mathcal{A}|}$ collects its action-conditional returns. The finite action set $\mathcal A$ contains at least two actions. For each action $a$, let $\mathbf{r}^{a}=(r_1(a),\ldots,r_n(a))^{\top}\in\mathbb{R}^{n}$ collect its returns across cases. Given a new situation $x$, CBDT assigns each previous case a nonnegative relevance weight $\omega_i(x)$, which quantifies how relevant case $i$ is to the current situation $x$ and evaluates action $a$ as $U(x,a)=\sum_{i=1}^{n}\omega_i(x)r_i(a)$, where $U(x,a)$ denotes the CBDT utility assigned to action $a$ under the current situation $x$.

For two candidate actions $a$ and $b$, define the return difference for case $i$ as $\Delta r_i(a,b)=r_i(a)-r_i(b).$ The preference difference can then be represented as
\setcounter{equation}{0}
\begin{equation}
\Delta U(x;a,b)=U(x,a)-U(x,b)=\sum_{i=1}^{n}\omega_i(x)\Delta r_i(a,b),\label{eq:1}
\end{equation}
where $\Delta U(x;a,b)$ denotes the preference for action $a$ over action $b$, and $\Delta r_i(a,b)$ represents the return difference between actions $a$ and $b$ for case $i$.

Equation~\eqref{eq:1} expresses decision preference as a sum of case contributions, where each contribution combines the relevance of a previous case with its relative return between competing actions. We refer to $\omega_i(x)\Delta r_i(a,b)$ as the case contribution of case $i$ to the preference between actions $a$ and $b$.

\subsection{Neural Network Decision Models}\label{neural-network-decision-models}

Neural networks learn complex decision functions from large collections of training cases through parameterized representations \citep{Bengio2013,LeCun2015}. Given training cases $x_1,\ldots,x_n$ and their task-specific supervision, a neural network learns a representation function
\setcounter{equation}{1}
\begin{equation}
\phi(x)=f_{\theta}(x)\in\mathbb{R}^{H},\label{eq:2}
\end{equation}
where $f_\theta$ denotes the neural network parameterized by $\theta$, and $\phi(x)$ denotes its learned $H$-dimensional representation.

The learned representation is used to produce scores for possible actions. For a set of candidate actions $\mathcal A$, the network assigns a score $S(x,a)$ to action $a$ through a scoring function $g_a(\cdot)$:
\setcounter{equation}{2}
\begin{equation}
S(x,a)=g_a(\phi(x)),\label{eq:3}
\end{equation}
The final decision is selected according to
\setcounter{equation}{3}
\begin{equation}
a_S(x)=\arg\max_{a\in\mathcal{A}}S(x,a),\label{eq:4}
\end{equation}
where $a_S(x)$ denotes the action selected by the neural network according to the score function $S$. All decision rules use the same deterministic rule to resolve ties.

Equation~\eqref{eq:3} maps an input through a learned representation to action scores \citep{Goodfellow2016}. Although the network learns from collections of cases, its final decision does not directly reveal how individual cases contribute to the selected action.

\subsection{Problem Setup}\label{problem-setup}

The two formulations above describe decision making from different perspectives. CBDT expresses decision preference through contributions from previous cases, whereas a neural network expresses it through differences between learned decision scores. Their relationship leads to the problem studied in this work.

Given two candidate actions $a$ and $b$, the neural network preference can be written as
\setcounter{equation}{4}
\begin{equation}
\Delta S(x;a,b)=S(x,a)-S(x,b),\label{eq:5}
\end{equation}
where $\Delta S(x;a,b)$ denotes the score difference between actions $a$ and $b$ for input $x$.

The problem considered in this work is whether the neural network preference in Equation~\eqref{eq:5} can be expressed through a case decomposition on its learned representation, following the case-based form of the CBDT preference decomposition in Equation~\eqref{eq:1}. We refer to the recoverable organization of these case-level contributions as case structure. The case coefficients $\alpha_i(x)$ introduced below are determined by the learned representation, and Section 4 will establish the conditions under which they obtain the CBDT interpretation of the nonnegative relevance weights $\omega_i(x)$ summarized in Table~\ref{tab:1}.

\begin{table}[tb]
\caption{Formal correspondence between CBDT decision decomposition and neural network case decomposition. The table illustrates how both formulations represent decision preferences through weighted case contributions, motivating the connection studied in this work. $\widehat{Q}$ is fitted on the representation.}\label{tab:1}
\begin{center}
\begin{tabular}{@{}
  >{\raggedright\arraybackslash}p{(\linewidth - 4\tabcolsep) * \real{0.2056}}
  >{\raggedright\arraybackslash}p{(\linewidth - 4\tabcolsep) * \real{0.3367}}
  >{\raggedright\arraybackslash}p{(\linewidth - 4\tabcolsep) * \real{0.4577}}@{}}
\toprule
\begin{minipage}[b]{\linewidth}\raggedright
Aspect
\end{minipage} & \begin{minipage}[b]{\linewidth}\raggedright
CBDT
\end{minipage} & \begin{minipage}[b]{\linewidth}\raggedright
Neural Network Case Decomposition
\end{minipage} \\
\midrule
Decision score & $U(x,a)=\sum_i \omega_i(x)r_i(a)$ & $\widehat{Q}(x,a)=\sum_i \alpha_i(x)r_i(a)$ \\
Case contribution & $\omega_i(x)r_i(a)$ & $\alpha_i(x)r_i(a)$ \\
Case coefficient & $\omega_i(x)\geq0$ & $\alpha_i(x)\in\mathbb{R}$ \\
Decision & $\arg\max_a U(x,a)$ & $\arg\max_a \widehat{Q}(x,a)$ \\
Interpretation & Case relevance under CBDT conditions & Case relevance when the conditions in Section 4 hold \\
\bottomrule
\end{tabular}
\end{center}
\end{table}

\section{Case Decomposition of Neural Network Decisions}\label{case-decomposition-of-neural-network-decisions}

A trained neural network representation provides the basis for expressing fitted decision scores in terms of individual training cases. To recover the case structure represented by these scores, Section 3.1 first formulates action-specific decision models over the learned representation, allowing Section 3.2 to identify the case coefficients shared across actions and obtain a case-level decomposition of the fitted scores. Building on this decomposition, Section 3.3 extends the case decomposition to decision margins between competing actions.

\subsection{Action-Specific Decision Models over a Learned Representation}\label{action-specific-decision-models-over-a-learned-representation}

To obtain a case decomposition, we first fit an action-specific return model over the learned representation of the neural network. Let $\phi(x)\in\mathbb R^H$ denote the learned representation of input $x$. For training cases $x_1,\ldots,x_n$, their representations are collected into the matrix
\setcounter{equation}{5}
\begin{equation}
\Phi=\begin{bmatrix}\phi(x_1)&\phi(x_2)&\cdots&\phi(x_n)\end{bmatrix}^{\top}\in\mathbb{R}^{n\times H}.\label{eq:6}
\end{equation}
The corresponding Gram matrix is $G=\Phi^\top\Phi\in\mathbb R^{H\times H}$.

For each candidate action $a\in\mathcal A$, we fit an action-specific linear return model on the learned representation:
\setcounter{equation}{6}
\begin{equation}
\widehat{Q}(x,a)=v_a^\top\phi(x),\label{eq:7}
\end{equation}
where $v_a\in\mathbb R^H$ is the ordinary least squares (OLS) coefficient minimizing $\left\Vert\Phi v-\mathbf r^a\right\Vert_2^2$. Here, $\mathbf r^a=(r_1(a),\ldots,r_n(a))^\top$ collects the returns of all training cases for action $a$, where $r_i(a)$ denotes the return associated with case $i$. Assuming $\operatorname{rank}(\Phi)=H$, which requires $n\geq H$, the solution is $v_a=G^{-1}\Phi^\top\mathbf r^a$. Substituting this solution into Eq.~\eqref{eq:7} gives
\setcounter{equation}{7}
\begin{equation}
\widehat{Q}(x,a)=\phi(x)^{\top}G^{-1}\Phi^{\top}\mathbf{r}^{a}.\label{eq:8}
\end{equation}
Equation~\eqref{eq:8} separates how the current input relates to the training cases through the learned representation from the action-specific returns. This relationship is shared across actions, while $\mathbf r^a$ changes with the action being evaluated. This fitted model is used only to expose the case structure induced by the learned representation of neural networks, rather than to replace the original neural network decision model.

\subsection{Case Decomposition of Decision Scores}\label{case-decomposition-of-decision-scores}

To decompose a fitted decision score into training-case contributions, we define a representation-based coefficient for each case. For training case $i$, define the case coefficient as
\setcounter{equation}{8}
\begin{equation}
\alpha_i(x)=\phi(x_i)^\top G^{-1}\phi(x).\label{eq:9}
\end{equation}
The coefficient $\alpha_i(x)$ measures the contribution assigned to case $i$ by the learned representation when reconstructing the decision score. It depends on the input representation and determines how each training case participates in the fitted decision function. Collecting all coefficients gives $\alpha(x)=(\alpha_1(x),\ldots,\alpha_n(x))^\top=\Phi G^{-1}\phi(x)$. Each coefficient depends on the current input and the learned representation, but not on the candidate action.

Substituting the case coefficients in Eq.~\eqref{eq:9} into the fitted score in Eq.~\eqref{eq:8} gives the following result.

\begin{proposition}[Shared case structure and decomposition]\label{prop:case-structure}
If $\operatorname{rank}(\Phi)=H$, the shared case coefficients in Eq.~\eqref{eq:9} give, for every action $a\in\mathcal A$,
\setcounter{equation}{9}
\begin{equation}
\widehat{Q}(x,a)=\sum_{i=1}^{n}\alpha_i(x)r_i(a).\label{eq:10}
\end{equation}
The coefficient vector $\alpha(x)$ is the unique minimum-Euclidean-norm solution satisfying $\Phi^\top\alpha(x)=\phi(x)$.
\end{proposition}
Equation~\eqref{eq:10} exposes the case structure defined in Section 2.3 by expressing the fitted score as a sum of training-case contributions. For case $i$, $\alpha_i(x)$ is shared across actions, while $r_i(a)$ provides the action-specific return. The same case structure therefore underlies the evaluation of different candidate actions.

Consequently, the case coefficients are determined by the learned representation rather than by a particular action or return vector. Proofs and the ridge extension are given in Appendices A.2--A.3.

\subsection{Decomposition of Decision Margins}\label{decomposition-of-decision-margins}

Decision selection depends on differences between the scores of candidate actions. To expose the case structure of these comparisons, we consider the fitted score difference between two candidate actions. For actions $a$ and $b$, define the fitted decision margin as
\setcounter{equation}{10}
\begin{equation}
\Delta\widehat{Q}(x;a,b)=\widehat{Q}(x,a)-\widehat{Q}(x,b).\label{eq:11}
\end{equation}
Using the case decomposition in Eq.~\eqref{eq:10}, and defining $\Delta r_i(a,b)=r_i(a)-r_i(b)$, this margin can be written as
\setcounter{equation}{11}
\begin{equation}
\Delta\widehat{Q}(x;a,b)=\sum_{i=1}^n\alpha_i(x)\Delta r_i(a,b).\label{eq:12}
\end{equation}
Equation~\eqref{eq:12} shows that each case enters the fitted margin through a representation-derived coefficient $\alpha_i(x)$ and an action-dependent return contrast $\Delta r_i(a,b)$. Their product defines the case-level contribution to that margin.

Accordingly, define the case-margin contribution of case $i$ as
\setcounter{equation}{12}
\begin{equation}
\chi_i(x;a,b)=\alpha_i(x)\Delta r_i(a,b).\label{eq:13}
\end{equation}
The complete fitted margin can then be decomposed as
\setcounter{equation}{13}
\begin{equation}
\Delta\widehat{Q}(x;a,b)=\sum_{i=1}^n\chi_i(x;a,b).\label{eq:14}
\end{equation}
The sign and magnitude of $\chi_i(x;a,b)$ show whether and how strongly case $i$ changes the fitted preference between actions $a$ and $b$. A positive contribution supports $a$ over $b$, whereas a negative contribution supports $b$ over $a$. Equation~\eqref{eq:10} decomposes each fitted action score into training-case terms, while Eq.~\eqref{eq:14} decomposes the corresponding pairwise margin. Whether this recovered case structure admits a Case-Based Decision Theory interpretation is addressed next.

\section{Case-Based Interpretation of Neural Network Decisions}\label{case-based-interpretation-of-neural-network-decisions}

Building on the recovered case structure in Section 3, this section establishes the conditions under which the recovered case contributions admit a CBDT interpretation.

\subsection{Conditions for Case-Based Interpretation}\label{conditions-for-case-based-interpretation}

The decomposition in Section 3 yields case-level coefficients, but these coefficients admit a Case-Based Decision Theory interpretation only when they satisfy the structural requirements of CBDT. In CBDT, decisions are formulated by combining the relevance of previous cases with their associated action returns. Specifically, the value of action $a$ is represented as $U(x,a)=\sum_{i=1}^{n}\omega_i(x)r_i(a)$.

The fitted decision score obtained from the neural network representation follows the same case-return structure: $\widehat Q(x,a)=\sum_{i=1}^n\alpha_i(x)r_i(a),$ where $\alpha_i(x)$ denotes the coefficient associated with case $i$. The two formulations differ in how these coefficients are obtained and interpreted. In CBDT, relevance weights $\omega_i(x)$ are nonnegative because they represent the relevance of previous cases when evaluating an action \citep{Gilboa1995}. In contrast, the neural network coefficients $\alpha_i(x)$ are determined by the learned representation and do not automatically satisfy the nonnegativity condition required for a CBDT interpretation. Therefore, the case coefficients admit a case-based interpretation when
\setcounter{equation}{14}
\begin{equation}
\alpha_i(x)\geq0,\qquad i=1,\ldots,n.\label{eq:15}
\end{equation}
Under this condition, $\alpha_i(x)$ can be interpreted as the CBDT relevance assigned to case $i$. The recovered case structure therefore admits a Case-Based Decision Theory interpretation in which each case contributes according to its relevance and associated return.

When this condition is not satisfied, the decomposition remains mathematically valid, but its coefficients cannot serve as CBDT relevance weights. Thus, case decomposition and Case-Based Decision Theory interpretation are distinct properties of the recovered case structure. A bound for clipping negative coefficients is derived in Appendix A.4, and coefficient signs and a decision example are examined in Appendix D.

\subsection{Decision Interpretation through Case Contributions}\label{decision-interpretation-through-case-contributions}

To interpret how individual cases enter a decision preference, we examine their contributions within a pairwise action comparison. For two candidate actions $a$ and $b$, the case contribution to the fitted decision margin is defined as $\chi_i(x;a,b)=\alpha_i(x)\Delta r_i(a,b)$. Under this interpretation condition, the coefficients $\alpha_i(x)$ can be assigned the CBDT meaning of relevance weights, allowing the recovered case contributions to represent supporting and opposing cases in the decision process.

The fitted decision margin is therefore the sum of individual case contributions:
\setcounter{equation}{15}
\begin{equation}
\Delta\widehat{Q}(x;a,b)=\sum_{i=1}^{n}\chi_i(x;a,b).\label{eq:16}
\end{equation}
A positive $\chi_i(x;a,b)$ supports $a$ over $b$, whereas a negative value supports $b$ over $a$. Under the interpretation condition, these contributions provide a case-based account of the fitted preference through its supporting and opposing cases.

\section{Decision Consistency of Case-Based Interpretations}\label{decision-consistency-of-case-based-interpretations}

The case-based interpretation describes the recovered case structure through the contributions of individual training cases. This section connects that interpretation to the original neural network by defining decision consistency and establishing a sufficient condition for matching decisions.

\subsection{Decision Consistency Criterion}\label{decision-consistency-criterion}

The case-based interpretation is obtained from the decomposed decision formulation $\widehat{Q}(x,a)$, while the original neural network produces decisions through its score function $S(x,a)$. Therefore, the interpretation corresponds to the original neural network decision when the fitted formulation and the neural network select the same action.

For an input $x$, the consistency requirement is defined as
\setcounter{equation}{16}
\begin{equation}
\displaystyle a_{\widehat{Q}}(x)=\arg\max_{a\in\mathcal{A}}\widehat{Q}(x,a)=\arg\max_{a\in\mathcal{A}}S(x,a)=a_S(x).\label{eq:17}
\end{equation}
where $a_{\widehat Q}(x)$ denotes the action selected by the fitted formulation and $a_S(x)$ denotes the action selected by the original neural network.

Decision consistency requires only that the fitted formulation and the neural network select the same action. It therefore addresses a different question from the CBDT interpretation condition in Section 4, which determines whether the case coefficients admit a case-based meaning.

\subsection{Sufficient Conditions for Decision Consistency}\label{sufficient-conditions-for-decision-consistency}

Decision consistency depends on whether the discrepancy between the fitted and neural network scores can change the selected action. Let the deviation between the original neural network score and the fitted decision score be bounded by
\setcounter{equation}{17}
\begin{equation}
\displaystyle |S(x,a)-\widehat{Q}(x,a)|\leq\varepsilon_{\mathrm{fid}}(x),\qquad a\in\mathcal{A}.\label{eq:18}
\end{equation}
where $\varepsilon_{\mathrm{fid}}(x)$ denotes the maximum deviation between the original neural network score and the fitted decision score for input $x$.

The following result links the fitted winning margin to decision consistency.

\begin{theorem}[Decision consistency]\label{thm:decision-consistency}
Fix an input $x$ and suppose Eq.~\eqref{eq:18} holds for every action. If the winning margin of the fitted formulation satisfies
\setcounter{equation}{18}
\begin{equation}
\displaystyle \widehat{Q}(x,a_{\widehat{Q}}(x))-\max_{b\neq a_{\widehat{Q}}(x)}\widehat{Q}(x,b)>2\varepsilon_{\mathrm{fid}}(x),\label{eq:19}
\end{equation}
then $a_{\widehat Q}(x)=a_S(x)$.
\end{theorem}
Under Eq.~\eqref{eq:19}, the score error cannot reverse the ordering between the fitted winner and any competitor, so the fitted formulation and the original neural network select the same action. The factor two accounts for score errors on both the fitted winner and a competitor. The proof is given in Appendix A.5.

Theorem~\ref{thm:decision-consistency} provides a sufficient condition for connecting the case-based interpretation of the recovered structure to the original neural network decision. Together with the CBDT interpretation condition in Section 4, it distinguishes case-based meaning from agreement with the neural network\textquotesingle s selected action.

\section{Experiments}\label{experiments}

We evaluate recoverable case structure in trained neural networks using a controlled CBDT setting and three decision tasks based on real-world data. The controlled setting provides explicit cases and decision outcomes for evaluating case decomposition (Section~\ref{case-decomposition-results}). The Pennsylvania--New Jersey--Maryland (PJM) bidding task evaluates case-based interpretation (Section~\ref{case-based-interpretation-results}) \citep{PJM2017}. Bidding, Adult Income \citep{Becker1996}, and Default of Credit Card Clients \citep{Yeh2009,YehLien2009} evaluate decision consistency with the original neural network (Section~\ref{decision-consistency-results}). Task construction, network configurations, and evaluation measures are specified in Appendix B.

\subsection{Case Decomposition Results}\label{case-decomposition-results}

We first evaluate whether the learned representation preserves a recoverable case structure by expressing fitted decision scores through individual case contributions.

\begin{figure}[!t]
\centering
\begin{tabular}{@{}c@{\hspace{8pt}}c@{\hspace{8pt}}c@{}}
\includegraphics[scale=0.41998529]{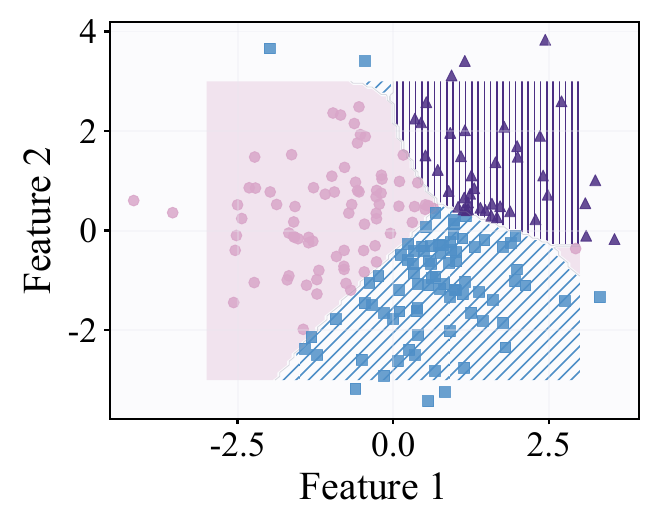}\llap{\raisebox{94bp}[0pt][0pt]{\makebox[130.027bp][l]{(a)}}} & \includegraphics[scale=0.41998529]{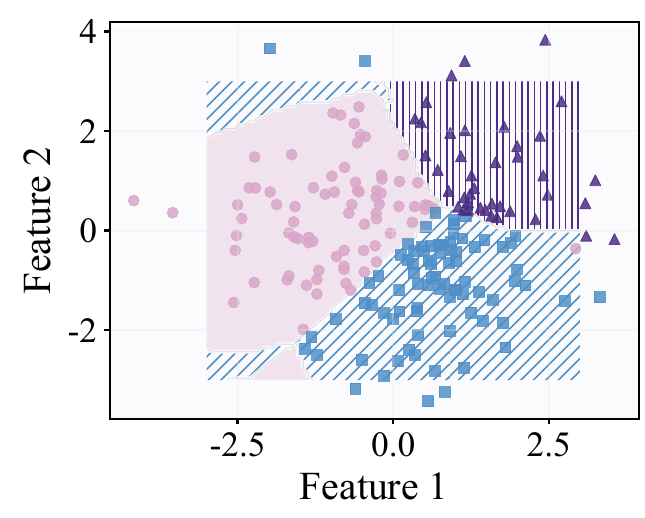}\llap{\raisebox{94bp}[0pt][0pt]{\makebox[130.027bp][l]{(b)}}} & \includegraphics[scale=0.41998529]{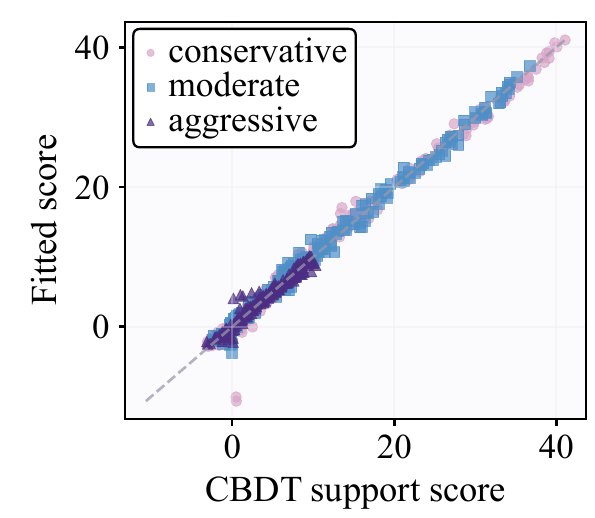}\llap{\raisebox{94bp}[0pt][0pt]{\makebox[119.946bp][l]{(c)}}}
\end{tabular}
\caption{Recovery of case structure from neural network representations on the controlled CBDT setting. (a) Ground-truth CBDT decision structure generated from the underlying case-based process. (b) Recovered decision structure obtained by decomposing fitted decision scores over the learned representation. The recovered decision regions closely align with the reference structure. (c) Correspondence between recovered and reference support scores, where the diagonal relationship indicates recovery of aggregate support.}\label{fig:2}
\end{figure}

\begin{table}[!t]
\caption{Case contribution correspondence with the reference structure on the controlled CBDT setting. Ranking is evaluated by Top-30 Jaccard and normalized discounted cumulative gain (nDCG); sign agreement measures support direction. The decomposition aligns more closely with the specified reference than the comparison scores.}\label{tab:2}
\begin{center}
\setlength{\tabcolsep}{4.5pt}
\begin{tabular}{@{}lcccc@{}}
\toprule
\textbf{Scoring rule} & \textbf{Pearson} $\uparrow$ &
\shortstack{\textbf{Top-30}\\\textbf{Jaccard} $\uparrow$} &
\shortstack{\textbf{Top-30}\\\textbf{nDCG} $\uparrow$} &
\shortstack{\textbf{Top-30}\\\textbf{sign agreement} $\uparrow$} \\
\midrule
\textbf{Case decomposition} & \textbf{0.600} & \textbf{0.390} & \textbf{0.711} & \textbf{0.828} \\
Influence Functions (linear model) & \ensuremath{-}0.023 & 0.124 & 0.360 & 0.498 \\
Representer-style (linear model) & 0.026 & 0.043 & 0.166 & 0.395 \\
Single-checkpoint gradient / inner product & \ensuremath{-}0.026 & 0.043 & 0.166 & 0.605 \\
\bottomrule
\end{tabular}
\end{center}
\end{table}

On the controlled CBDT setting, the recovered decision structure closely matches the original case-based structure (Figure~\ref{fig:2}(a,b)), showing that the learned representation preserves the organization required for case-level decomposition. The recovered contributions further preserve the relative importance and support direction of individual cases (Figure~\ref{fig:2}(c); Table~\ref{tab:2}), indicating that the decomposed case-level contributions capture the decision organization represented by the learned model. The representation ablation further examines the role of the learned representation: the trained representation reconstructs neural network scores with $R^2=1.000$, produces support correlation of 0.996, and satisfies the certification condition for 94.3\% of evaluated inputs, while replacing it with random representations substantially reduces recovery consistency. These results indicate that the recoverable case structure is supported by information encoded in the learned representation. Full ablation and checkpoint results are reported in Appendix C, with comparison details and costs in Appendix F.

\subsection{Case-Based Interpretation Results}\label{case-based-interpretation-results}

We next examine whether the case decomposition obtained from neural network decisions admits a Case-Based Decision Theory interpretation.

\begin{figure}[!t]
\centering
\begin{tabular}{@{}c@{\hspace{8pt}}c@{}}
\includegraphics[scale=0.53465942]{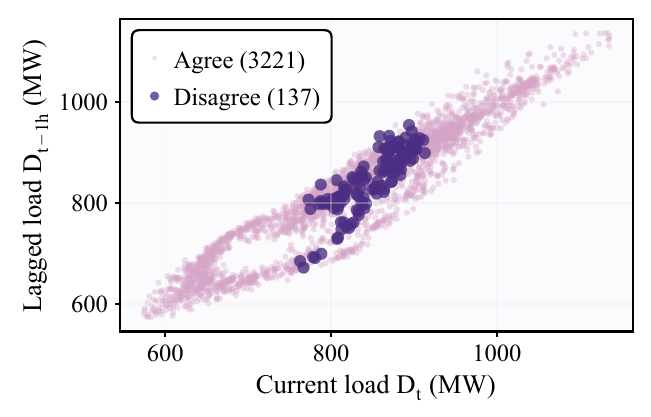}\llap{\raisebox{98.917bp}[0pt][0pt]{\makebox[167.071bp][l]{(a)}}} & \includegraphics[scale=0.53465942]{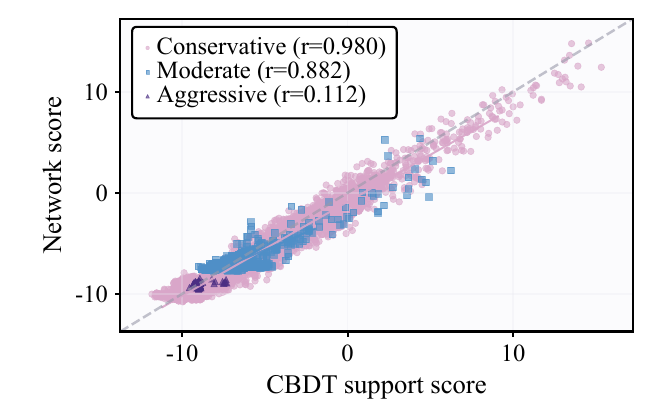}\llap{\raisebox{98.917bp}[0pt][0pt]{\makebox[167.071bp][l]{(b)}}}
\end{tabular}
\caption{Decision agreement and case support correspondence on the PJM bidding task. (a) Agreement between neural network and CBDT selected actions, where the two models select the same bidding action for 95.9\% of inputs; loads are in megawatts (MW). (b) Conservative, Moderate, and Aggressive denote the three bidding strategies evaluated in the task. Correspondence between neural network scores and explicit CBDT support, where the Aggressive bidding action achieves only 0.112 support-score correlation despite action agreement.}\label{fig:3}
\end{figure}

On the PJM bidding task, we find that decision agreement alone is insufficient to identify a case-based interpretation. The comparison in Figure~\ref{fig:3} shows that the neural network and the CBDT model can achieve highly consistent decisions while exhibiting substantially different support patterns. This observation distinguishes action-level consistency from case-based interpretation and motivates the interpretation condition introduced in Section 4. We further evaluate this condition on recovered case decompositions. Across 250 bidding inputs, the decomposition remains mathematically valid, but every coefficient vector contains negative entries and therefore fails the direct CBDT interpretation condition. Detailed condition analysis and additional decision records are reported in Appendix D.

\begin{figure}[!t]
\centering
\begin{tabular}{@{}c@{\hspace{8pt}}c@{}}
\includegraphics[scale=0.76733528]{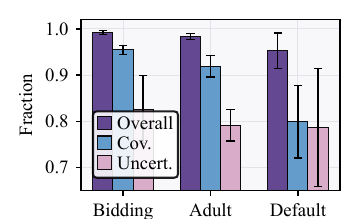}\llap{\raisebox{71.662bp}[0pt][0pt]{\makebox[127.071bp][l]{(a)}}} & \includegraphics[scale=0.76733528]{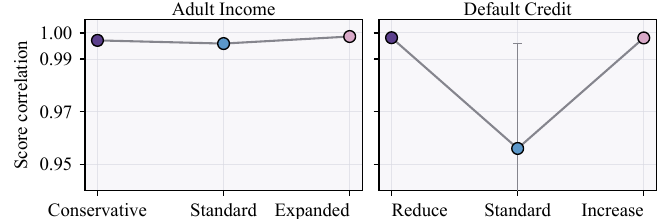}\llap{\raisebox{71.662bp}[0pt][0pt]{\makebox[244.197bp][l]{(b)}}}
\end{tabular}
\caption{Decision consistency across the three tasks. (a) Overall agreement, certification coverage (Cov.), and agreement on inputs without certification (Uncert.), where Cov. denotes the fraction of inputs satisfying the decision consistency condition. Coverage is 95.5\% for bidding, 91.9\% for Adult Income, and 79.9\% for Default Credit. (b) Fitted--network score correlation by action; Adult correlations are nearly identical, whereas Default Credit varies across actions.}\label{fig:4}
\end{figure}

\subsection{Decision Consistency Results}\label{decision-consistency-results}

We finally examine whether the case structure preserved in neural network decisions supports decisions consistent with the original models. This evaluation completes the connection from case decomposition and interpretation to decision consistency.

Across the three decision tasks, the fitted decisions largely agree with the original neural network decisions (Figure~\ref{fig:4}). These consistency results show that the recovered case structure preserves the decision preferences represented by the original model beyond individual case contributions. The preference correspondence across actions further indicates that the recovered structure can maintain the original decision rule while allowing action-specific variation across tasks. These results connect the recovered case structure with the original neural network decisions while preserving the distinction between case-based interpretation and decision consistency. Additional seed-level and sensitivity analyses are reported in Appendix E.

\section{Conclusion}\label{conclusion}

Neural networks increasingly inform high-impact decisions, yet how training cases shape their choices remains unclear. We expose recoverable case structure in learned representations through case-level decomposition, characterize when it admits a Case-Based Decision Theory interpretation, and establish when that interpretation is consistent with the original network decision. These results connect final choices with prior cases, providing a unified view of how learned decision systems accumulate case information. Future work will investigate more diverse neural architectures and decision settings.

\bibliographystyle{plainnat}
\bibliography{references}

\clearpage
\appendix
\renewcommand{\theequation}{\thesection\arabic{equation}}
\renewcommand{\thetable}{\thesection\arabic{table}}
\renewcommand{\thefigure}{\thesection\arabic{figure}}
\section{Mathematical Details}\label{a-mathematical-details}
\setcounter{table}{0}
\setcounter{figure}{0}

\begingroup
\setlength{\LTcapwidth}{\linewidth}
\begin{longtable}{@{}
  >{\raggedright\arraybackslash}p{(\linewidth - 6\tabcolsep) * \real{0.1800}}
  >{\raggedright\arraybackslash}p{(\linewidth - 6\tabcolsep) * \real{0.3200}}
  >{\raggedright\arraybackslash}p{(\linewidth - 6\tabcolsep) * \real{0.1800}}
  >{\raggedright\arraybackslash}p{(\linewidth - 6\tabcolsep) * \real{0.3200}}@{}}
\caption{Notation used in the case decomposition, CBDT interpretation condition, decision-consistency bound, and experimental evaluations.}\label{tab:A1}\\
\toprule
\begin{minipage}[b]{\linewidth}\raggedright
\textbf{Symbol}
\end{minipage} & \begin{minipage}[b]{\linewidth}\raggedright
\textbf{Meaning}
\end{minipage} & \begin{minipage}[b]{\linewidth}\raggedright
\textbf{Symbol}
\end{minipage} & \begin{minipage}[b]{\linewidth}\raggedright
\textbf{Meaning}
\end{minipage} \\
\midrule\endfirsthead
\toprule
\textbf{Symbol} & \textbf{Meaning} & \textbf{Symbol} & \textbf{Meaning} \\
\midrule\endhead
$\mathcal{X},\mathcal{A}$ & Input space and finite action set & $\mathcal{D}$ & Collection of training cases and returns \\
$x,\ x_i$ & Query and training input $i$ & $n,\ H$ & Number of training cases; representation dimension \\
$a,\ b$ & Two candidate actions & $\mathbf{r}_i$ & Returns for case $i$; dimension $|\mathcal{A}|$ \\
$r_i(a)$ & Return of action $a$ for case $i$ & $\mathbf{r}^{a}$ & Returns across $n$ cases for action $a$ \\
$\phi(x)$ & Column representation; dimension $H$ & $\Phi$ & Training representation matrix; $n\times H$ \\
$G$ & Gram matrix $\Phi^{\top}\Phi$; $H\times H$ & $I_H$ & Identity matrix of dimension $H$ \\
$S(x,a)$ & Neural network score for action $a$ & $\widehat{Q}(x,a)$ & Fitted score; OLS in Section 3, chosen fit in Sections 4--5 \\
$v_a$ & OLS weight vector for action $a$ & $\alpha_i(x)$ & Coefficient of case $i$ in the chosen fit \\
$\alpha(x)$ & Action-shared coefficient vector; dimension $n$ & $\lambda$ & Ridge regularization strength \\
$\alpha_{\lambda}(x)$ & Ridge vector; entry $\alpha_{\lambda,i}(x)$ & $\widehat{Q}_\lambda$ & Ridge-fitted return score \\
$\Delta r_i(a,b)$ & Case return contrast: $r_i(a)-r_i(b)$ & $\chi_i(x;a,b)$ & Pairwise case contribution; $\chi_{\lambda,i}$ for ridge \\
$\Delta\widehat{Q}$ & Fitted score difference between actions & $\omega_i(x)$ & Nonnegative CBDT case relevance \\
$U(x,a)$ & CBDT support for action $a$ & $a_S(x)$ & Action selected by the neural network \\
$a_{\widehat{Q}}(x)$ & Action selected by the fitted model & $m_{\widehat{Q}}(x)$ & Fitted winning margin \\
$\varepsilon_{\mathrm{fid}}(x)$ & Maximum fitted--network score error & $\alpha_i^+(x),\alpha_i^-(x)$ & Positive and negative coefficient parts \\
$\nu(x),\nu_\lambda(x)$ & Negative mass for the chosen fit and ridge & $R_a$ & Maximum absolute return for action $a$ \\
$Q^+(x,a)$ & Score after clipping coefficients at zero & $k,\ \Phi_k$ & Checkpoint index and its representation matrix \\
$G_k,\alpha_{k,\lambda}$ & Checkpoint Gram matrix and ridge coefficients & $B,\ G_B$ & Invertible coordinate map and transformed Gram matrix \\
$a_{\mathrm{gt}},b_{\mathrm{gt}}$ & Generator-selected action and runner-up & $\chi_i^{\mathrm{gt}}$ & Reference contribution on sampled training cases \\
$\sigma_r,\sigma_a,\eta_{i,a}$ & Noise level; return standard deviation; Gaussian draw & $R^2$, error & Coefficient of determination; mean squared error (MSE) \\
Jaccard, nDCG & Case-set overlap and normalized discounted cumulative gain & Uncertainty & Standard deviation (SD); confidence interval (CI) \\
$f_\theta,g_a$ & Representation map and action scoring function & $\theta$ & Neural network parameters \\
$v_{\lambda,a}$ & Ridge weight vector for action $a$ & $\Delta S,\Delta U$ & Network and CBDT score differences \\
$\phi_k,\widehat{Q}_{k,\lambda}$ & Checkpoint representation and fitted score & $\widehat{a}$ & Shorthand for $a_{\widehat{Q}}(x)$ \\
$N,\bar S_a$ & Evaluation count and mean network action score & $\tilde r_i(a),\bar r_a$ & Perturbed return and mean training return \\
$K_i(x),e_i(a,b)$ & Feature inner product and residual contrast & $H_d,\lambda_{\mathrm{rep}}$ & Damped comparison matrix and representer scale \\
$E_{\lambda}^{\mathrm{avg}}$ & Action-averaged clipping bound & $A_{\mathrm{all}},A_{\mathrm{unc}}$ & Overall and uncertified agreement rates \\
\bottomrule
\end{longtable}
\endgroup

Sections 3--5 address three properties of the recovered case structure: how a learned representation yields a case decomposition, when the recovered coefficients admit a case-based interpretation, and when the resulting case-based formulation remains consistent with the original neural network decision. The notation is summarized in Table~\ref{tab:A1}, followed by the proofs in that order. Experimental settings are specified in Appendix B, with corresponding evidence in Appendices C--E and complementary comparisons and computational costs in Appendix F.

\subsection{Notation and assumptions}\label{a.1-notation-and-assumptions}

Let $\mathcal{D}=\{(x_i,\mathbf{r}_i)\}_{i=1}^{n}$ be the training cases, with $\mathbf{r}_i\in\mathbb{R}^{|\mathcal{A}|}$ and $\mathbf{r}^{a}\in\mathbb{R}^{n}$ as defined in Section 2.1. The action set is finite and contains at least two actions. A common deterministic rule resolves ties. The query $x\in\mathcal{X}$ may be a training input or a new input, where $\mathcal{X}$ denotes the input space. Returns are fixed when fitting the linear return models; the experiments construct a return for every action and case.

The neural network supplies scores $S(x,a)$ and a learned representation $\phi(x)\in\mathbb{R}^{H}$. During case-structure recovery, this learned representation is treated as fixed. Row $i$ of $\Phi\in\mathbb{R}^{n\times H}$ is $\phi(x_i)^{\top}$, and $G=\Phi^{\top}\Phi\in\mathbb{R}^{H\times H}$. The OLS results assume $\operatorname{rank}(\Phi)=H$, so $n\geq H$ and $G$ is positive definite. Ridge fitting uses $\lambda>0$ and also applies when this rank condition fails. All actions share the same cases, representation, fitting objective, and regularization strength.

We use $\widehat{Q}$ and $\alpha$ for OLS in Section 3 and Appendix A.2, and write $\widehat{Q}_{\lambda}$ and $\alpha_{\lambda}$ when specifying ridge fitting. In Sections 4--5 and Appendices A.4--A.5, $\widehat{Q}$ denotes the chosen fitted score and $\alpha$ its case coefficients. The identities decompose this fitted score; Eq.~\eqref{eq:19} connects its selected action to the network. The experiments retain the network output bias in $S$ and fit the return models without an additional intercept.

\subsection{Case decomposition and the minimum-norm coefficient vector (Proposition~\ref{prop:case-structure})}\label{a.2-case-decomposition-and-the-minimum-norm-coefficient-vector}

The shared representation is what allows all actions to use one case coefficient vector. For action $a$, differentiate the squared fitting loss:
\setcounter{equation}{0}
\begin{equation}
\nabla_v\left\Vert\Phi v-\mathbf{r}^{a}\right\Vert_2^2=2\Phi^{\top}(\Phi v-\mathbf{r}^{a}).\label{eq:A1}
\end{equation}
The normal equation is $Gv_a=\Phi^\top\mathbf{r}^{a}$. Full column rank makes $G$ invertible, so substitution into the fitted score gives
\setcounter{equation}{1}
\begin{equation}
\widehat{Q}(x,a)=\phi(x)^\top G^{-1}\Phi^\top\mathbf{r}^{a}=\alpha(x)^\top\mathbf{r}^{a},\label{eq:A2}
\end{equation}
where $\alpha(x)=\Phi G^{-1}\phi(x)$. Its $i$th entry is the coefficient in Eq.~\eqref{eq:9}, since $G^{-1}$ is symmetric. Neither the action index nor the return vector appears in $\alpha(x)$. Equation~\eqref{eq:A2} therefore establishes Eq.~\eqref{eq:10} simultaneously for every action.

To establish the characterization in Section 3.2, consider
\setcounter{equation}{2}
\begin{equation}
\min_{\beta\in\mathbb{R}^{n}}\left\Vert\beta\right\Vert_2^2\quad\mathrm{subject\ to}\quad\Phi^{\top}\beta=\phi(x).\label{eq:A3}
\end{equation}
The proposed vector is feasible because $\Phi^\top\alpha(x)=GG^{-1}\phi(x)=\phi(x)$. Every other feasible vector has the form $\beta=\alpha(x)+z$ with $\Phi^\top z=0$. Since $\alpha(x)$ belongs to the column space of $\Phi$, $\alpha(x)^\top z=\phi(x)^\top G^{-1}\Phi^\top z=0$. Expanding the squared norm therefore gives
\setcounter{equation}{3}
\begin{equation}
\left\Vert\beta\right\Vert_2^2=\left\Vert\alpha(x)\right\Vert_2^2+\left\Vert z\right\Vert_2^2.\label{eq:A4}
\end{equation}
Hence $\|\beta\|_2^2\geq\|\alpha(x)\|_2^2$, with equality only for $z=0$. This proves that $\alpha(x)$ is the unique minimum-norm vector satisfying the representation constraint. The uniqueness concerns this specified representation and OLS objective. It does not require the training returns to determine a unique expansion by themselves.

For any two actions, subtracting their score decompositions yields
\setcounter{equation}{4}
\begin{equation}
\Delta\widehat{Q}(x;a,b)=\sum_{i=1}^n\alpha_i(x)\Delta r_i(a,b)=\sum_{i=1}^n\chi_i(x;a,b).\label{eq:A5}
\end{equation}
This proves Eqs. (12)--(14). All training cases enter the identity. A displayed list of supporting and opposing cases is a partial view of that sum.

\subsection{Ridge fitting and checkpoint-specific case maps}\label{a.3-ridge-fitting-and-checkpoint-specific-case-maps}

The experiments use ridge fitting with $\lambda>0$. For each action, $v_{\lambda,a}$ minimizes $\left\Vert\Phi v-\mathbf{r}^{a}\right\Vert_2^2+\lambda\left\Vert v\right\Vert_2^2$. The Hessian is $2(G+\lambda I_H)$, and $z^\top(G+\lambda I_H)z=\|\Phi z\|_2^2+\lambda\|z\|_2^2>0$ for every nonzero $z\in\mathbb{R}^H$. Thus the objective has a unique minimizer, including when $\Phi$ is rank deficient. Its normal equation is $(G+\lambda I_H)v_{\lambda,a}=\Phi^{\top}\mathbf{r}^{a}$, giving
\setcounter{equation}{5}
\begin{equation}
v_{\lambda,a}=(G+\lambda I_H)^{-1}\Phi^{\top}\mathbf{r}^{a}.\label{eq:A6}
\end{equation}
\setcounter{equation}{6}
\begin{equation}
\alpha_\lambda(x)=\Phi(G+\lambda I_H)^{-1}\phi(x).\label{eq:A7}
\end{equation}
The fitted ridge score is $\widehat{Q}_{\lambda}(x,a)=v_{\lambda,a}^{\top}\phi(x)=\alpha_{\lambda}(x)^{\top}\mathbf{r}^{a}$. Its pairwise margin therefore has contributions $\chi_{\lambda,i}(x;a,b)=\alpha_{\lambda,i}(x)\Delta r_i(a,b)$. Both expansions are exact for this fitted model. The coefficients satisfy
\setcounter{equation}{7}
\begin{equation}
\Phi^{\top}\alpha_{\lambda}(x)=\phi(x)-\lambda(G+\lambda I_H)^{-1}\phi(x).\label{eq:A8}
\end{equation}
Equation~\eqref{eq:A8} follows from $\Phi^\top\alpha_\lambda=G(G+\lambda I_H)^{-1}\phi$ and $G=(G+\lambda I_H)-\lambda I_H$. The residual distinguishes ridge fitting from the equality constraint in Eq.~\eqref{eq:A3}. When $G$ is invertible, $\alpha_{\lambda}(x)$ converges to the OLS vector $\alpha(x)$ as $\lambda\to0$.

The same construction applies at any saved checkpoint $k$. Holding that checkpoint\textquotesingle s representation fixed gives
\setcounter{equation}{8}
\begin{equation}
\alpha_{k,\lambda}(x)=\Phi_k(G_k+\lambda I_H)^{-1}\phi_k(x),\label{eq:A9}
\end{equation}
Here $G_k=\Phi_k^{\top}\Phi_k$, and $\widehat{Q}_{k,\lambda}(x,a)=\alpha_{k,\lambda}(x)^{\top}\mathbf{r}^{a}$. For OLS, set $\lambda=0$ under full column rank. Each checkpoint receives newly fitted return-model weights. The resulting changes in the case coefficients are examined in Appendix C.2.

\subsection{Conditions for Case-based interpretation and the Nonnegative Approximation}\label{a.4-conditions-for-case-based-interpretation-and-the-nonnegative-approximation}

For either fitted model, nonnegative coefficients $\alpha_i(x)\geq0$ can serve as CBDT relevance weights by setting $\omega_i(x)=\alpha_i(x)$. This choice reproduces every fitted score and pairwise margin. The formulation uses nonnegative weights without a unit-sum constraint; multiplying all weights by a common positive number preserves the selected action. A negative entry prevents that coefficient vector from serving directly as nonnegative relevance weights, while leaving its signed decomposition intact.

Clipping replaces negative coefficients by zero. Define $\alpha_i^{+}(x)=\max\{\alpha_i(x),0\}$ and $\alpha_i^{-}(x)=\max\{-\alpha_i(x),0\}$, so $\alpha_i(x)=\alpha_i^{+}(x)-\alpha_i^{-}(x)$. Let $\nu(x)=\sum_{i=1}^{n}\alpha_i^{-}(x)$ and $Q^{+}(x,a)=\sum_{i=1}^{n}\alpha_i^{+}(x)r_i(a)$. If $|r_i(a)|\leq R_a$ for every case, then
\setcounter{equation}{9}
\begin{equation}
|\widehat{Q}(x,a)-Q^{+}(x,a)|=\left|\sum_{i=1}^{n}\alpha_i^{-}(x)r_i(a)\right|\leq\nu(x)R_a.\label{eq:A10}
\end{equation}
Cancelling the positive parts gives $\widehat{Q}-Q^+=-\sum_i\alpha_i^-r_i(a)$. Taking absolute values removes the leading minus sign, and the triangle inequality gives $|\sum_i\alpha_i^-r_i(a)|\leq\sum_i\alpha_i^-|r_i(a)|\leq\nu R_a$. In the ridge experiments, the negative mass is $\nu_{\lambda}(x)=\sum_{i=1}^{n}\max\{-\alpha_{\lambda,i}(x),0\}$.

For the pairwise clipped margin $\Delta Q^{+}(x;a,b)=Q^{+}(x,a)-Q^{+}(x,b)$, the same argument gives
\setcounter{equation}{10}
\begin{equation}
|\Delta\widehat{Q}(x;a,b)-\Delta Q^{+}(x;a,b)|\leq\nu(x)\max_i|\Delta r_i(a,b)|.\label{eq:A11}
\end{equation}
Since $\max_i|\Delta r_i(a,b)|\leq R_a+R_b$, the action-wise return bounds also control this difference. A small bound supports a close nonnegative approximation on the chosen return scale. A large bound leaves that approximation unresolved; it is not a measurement of the actual clipping error. The observed scale on the bidding proxy is reported in Appendix D.

If the sum of the clipped coefficients is positive, dividing all of them by that sum preserves the action selected by $Q^+$. This normalization rescales its scores, so Eq.~\eqref{eq:A10} refers to direct clipping before normalization. The same distinction applies to ridge coefficients.

\subsection{Sufficient condition for decision consistency (Theorem~\ref{thm:decision-consistency})}\label{a.5-sufficient-condition-for-decision-consistency}

Fix either fitted model and let $\widehat{a}=a_{\widehat{Q}}(x)$ be its selected action at the query under consideration. Its winning margin is
\setcounter{equation}{11}
\begin{equation}
m_{\widehat{Q}}(x)=\widehat{Q}(x,\widehat{a})-\max_{b\in\mathcal{A},\ b\neq\widehat{a}}\widehat{Q}(x,b).\label{eq:A12}
\end{equation}
Set $\varepsilon_{\mathrm{fid}}(x)=\max_{a\in\mathcal{A}}|S(x,a)-\widehat{Q}(x,a)|$. Writing $\delta_a=S(x,a)-\widehat{Q}(x,a)$ gives $|\delta_a|\leq\varepsilon_{\mathrm{fid}}(x)$ for every action. For any competitor $b\neq\widehat{a}$,
\setcounter{equation}{12}
\begin{equation}
\begin{aligned}
S(x,\widehat{a})-S(x,b)
&=\widehat{Q}(x,\widehat{a})-\widehat{Q}(x,b)+\delta_{\widehat{a}}-\delta_b\\
&\geq\widehat{Q}(x,\widehat{a})-\widehat{Q}(x,b)-|\delta_{\widehat{a}}|-|\delta_b|\\
&\geq\widehat{Q}(x,\widehat{a})-\widehat{Q}(x,b)-2\varepsilon_{\mathrm{fid}}(x).
\end{aligned}\label{eq:A13}
\end{equation}
The right-hand side is at least $m_{\widehat{Q}}(x)-2\varepsilon_{\mathrm{fid}}(x)$. Under Eq.~\eqref{eq:19}, it is strictly positive for every competitor. Hence $a_S(x)=\widehat{a}=a_{\widehat{Q}}(x)$. The proof uses the fitted scores and their error, so it applies to both OLS and ridge.

The factor two accounts for errors of opposite signs on the selected action and its competitor. Decision consistency establishes the same maximizing action. It leaves the ordering among the remaining actions unrestricted, and it does not equate the numerical margins of the two models. The strict condition can fail even when their selected actions agree. Coverage of the sufficient condition and observed agreement outside that set are evaluated in Appendix E.

This result is independent of coefficient nonnegativity. A signed case decomposition can have a certified matching decision. A CBDT interpretation of that decision requires both the nonnegativity condition in Section 4 and decision consistency.

\subsection{Coordinate changes and score scale}\label{a.6-coordinate-changes-and-score-scale}

For OLS under full column rank, let $B\in\mathbb{R}^{H\times H}$ be invertible and define $\phi_B(x)=B\phi(x)$. Since training representations are rows, $\Phi_B=\Phi B^{\top}$ and $G_B=BGB^{\top}$. Therefore
\setcounter{equation}{13}
\begin{equation}
\Phi_B G_B^{-1}\phi_B(x)=\Phi B^{\top}B^{-\top}G^{-1}B^{-1}B\phi(x)=\alpha(x).\label{eq:A14}
\end{equation}
Thus the OLS coefficients, their signs, and their case-margin contributions are unchanged. If the uncentered second moment is whitened so that $G=nI_H$, then $\alpha_i(x)=\phi(x)^{\top}\phi(x_i)/n$. This inner product can be negative; whitening does not imply nonnegative relevance.

With ridge fitting, an invertible coordinate change can alter the coefficients when the penalty remains $\lambda I_H$. For orthogonal $B$, $G_B+\lambda I_H=B(G+\lambda I_H)B^\top$ because $BB^\top=I_H$. Substitution into Eq.~\eqref{eq:A7} then gives the same ridge coefficients.

Decision consistency is also invariant to a common positive score scale. Replacing both score systems by $S'(x,a)=cS(x,a)+d(x)$ and $\widehat{Q}'(x,a)=c\widehat{Q}(x,a)+d(x)$, with $c>0$ and an action-independent offset $d(x)$, multiplies both $m_{\widehat{Q}}$ and $\varepsilon_{\mathrm{fid}}$ by $c$. Equation~\eqref{eq:19} is unchanged.

\section{Experimental Setting and Evaluation}\label{b-experimental-setting-and-evaluation}
\setcounter{table}{0}
\setcounter{figure}{0}

The experiments evaluate three properties of the recovered case structure: correspondence with known case contributions, case-based interpretation, and consistency with the original network decision. The controlled setting supplies a known support function. Adult, Default Credit, and the bidding task use real-world decision inputs and constructed action returns. Each case therefore has the complete return vector required by the analysis.

\subsection{Tasks and fitted models}\label{b.1-tasks-and-fitted-models}

The controlled environment contains 20 latent CBDT memory cases in two dimensions, three actions, and Gaussian similarity of bandwidth 1. The memory locations and action-return parameters are sampled once per seed. Their weighted support function generates returns for 500 training inputs and 200 test inputs. The 500 sampled training problems form the decomposition\textquotesingle s case collection; the 20 latent cases define the environment. This setting tests agreement with a known support function and a specified margin reference; the two case collections are distinct.

Adult uses six numeric covariates from 30,162 complete records of the University of California, Irvine (UCI) dataset \citep{Becker1996}. Default uses 23 coded covariates from 30,000 records in Default of Credit Card Clients \citep{Yeh2009,YehLien2009}. Both use seed-specific random 80/20 splits. The bidding proxy uses 2017 PJM load data \citep{PJM2017}. It contains 365 ordered days with 48 slots per day; the task treats each slot as half an hour. Current load, one-hour-lagged load, and cyclic time features form each input. Dropping the first two slots of each day gives 16,790 instances, split chronologically 80/20. Network configurations are listed in Table~\ref{tab:B1}.

\begin{table}[htbp]
\caption{Network configurations and data splits for the controlled setting and real-data tasks. All networks have three action outputs. The wider PJM network is used for coefficient-sign analysis and the individual decision example; the narrower network is used for cross-task consistency. Train/test counts refer to the task inputs after preprocessing.}\label{tab:B1}
\begin{center}
\begin{tabular}{@{}
  >{\raggedright\arraybackslash}p{(\linewidth - 6\tabcolsep) * \real{0.3182}}
  >{\centering\arraybackslash}p{(\linewidth - 6\tabcolsep) * \real{0.2000}}
  >{\centering\arraybackslash}p{(\linewidth - 6\tabcolsep) * \real{0.1818}}
  >{\centering\arraybackslash}p{(\linewidth - 6\tabcolsep) * \real{0.3000}}@{}}
\toprule
\begin{minipage}[b]{\linewidth}\raggedright
\textbf{Experiment / task}
\end{minipage} & \begin{minipage}[b]{\linewidth}\centering
\textbf{Hidden widths}
\end{minipage} & \begin{minipage}[b]{\linewidth}\centering
\textbf{Max. iterations}
\end{minipage} & \begin{minipage}[b]{\linewidth}\centering
\textbf{Train / test}
\end{minipage} \\
\midrule
Controlled CBDT & 32, 16 & 2,000 & 500 / 200 \\
PJM consistency & 32, 16 & 2,000 & 13,432 / 3,358 \\
PJM signs and case example & 128, 64, 32 & 2,000 & 13,432 / 3,358 \\
Adult & 64, 32, 16 & 2,000 & 24,129 / 6,033 \\
Default & 64, 32, 16 & 800 & 24,000 / 6,000 \\
\bottomrule
\end{tabular}
\end{center}
\end{table}

Every network is a multilayer perceptron (MLP) with rectified linear unit (ReLU) hidden layers and three linear outputs and is trained on squared return error with adaptive moment estimation (Adam). The initial learning rate is $10^{-3}$ and weight regularization is $10^{-4}$. Input standardization uses the training partition. Default selects a stopping point using a 10\% validation subset and patience 20; the other final-model runs use training-loss stopping criteria within their stated iteration limits.

After training, the final hidden representation is fixed. Each action receives a ridge return model with $\lambda=10^{-5}$ and no additional intercept. All actions share the same representation, case collection, and regularization, so their recovered case coefficients can be compared through the action-specific return contrasts.

\subsection{What the constructed returns represent}\label{b.2-what-the-constructed-returns-represent}

Adult\textquotesingle s actions represent conservative, standard, and expanded approval. Their base returns depend on income, education, and capital gains, with fixed Gaussian perturbations of standard deviation 0.05 applied to all action-return entries. Default\textquotesingle s reduce, maintain, and increase actions use a constructed risk score combining the observed default label, repayment delays, credit limit, and repayment-to-bill ratio. Its returns include fixed Gaussian perturbations of standard deviation 0.04. The maintain action is labeled Standard in the reported figures and tables. The normalization used to define Default\textquotesingle s utility is computed before splitting; network-input normalization is fitted on the training partition.

The bidding task attaches returns to low, medium, and high bids at 25, 45, and 70. A load-dependent synthetic clearing price determines whether a bid is accepted and its resulting return. The real data supply the covariates in these tasks; the action utilities are defined by the experiment. The stored return tables and construction code specify these utilities.

The explicit CBDT rule in Figure~\ref{fig:3} provides a reference case-based decision process by combining the returns of the 50 most similar training cases with nonnegative Gaussian input-based weights. This comparison evaluates whether neural network decisions and case-based decisions can exhibit similar outcomes while relying on different case contribution patterns. The decision-consistency experiments compare the recovered case-based formulation $\widehat{Q}_\lambda$ with the original network scores $S$.

\subsection{Reference contributions and distinct evaluation quantities}\label{b.3-reference-contributions-and-distinct-evaluation-quantities}

The controlled margin reference uses the generator-selected action $a_{\mathrm{gt}}$ and runner-up $b_{\mathrm{gt}}$. For each of the 500 sampled training cases, it is
\setcounter{equation}{0}
\begin{equation}
\chi_i^{\mathrm{gt}}(x)=\exp(-\|x-x_i\|_2^2/2)\,[r_i(a_{\mathrm{gt}})-r_i(b_{\mathrm{gt}})].\label{eq:B1}
\end{equation}
Every compared method uses this action pair. Equation~\eqref{eq:B1} defines a Gaussian relevance-and-return reference over the 500 sampled training cases. The generating support function instead uses 20 latent memory cases. The comparison measures alignment with the specified reference over training cases.

Pearson correlation compares signed scores across all training cases. Top-30 Jaccard compares the case sets with the largest absolute method and reference scores. Normalized discounted cumulative gain (nDCG) uses absolute reference contributions as gains. The signed Top-30 measure is direction agreement at the method\textquotesingle s 30 retrieved indices; it does not also require those indices to belong to the reference Top-30 set.

Two additional case statistics answer different questions. In the representation ablation, Top-30 action consistency uses the $|\alpha_\lambda|$-weighted majority action of the 30 largest absolute coefficients and compares it with the generator\textquotesingle s selected action. In the checkpoint analysis, Jaccard compares the Top-30 absolute-coefficient sets at the current and final checkpoints.

Reconstruction $R^2$ measures how closely the fitted scores reproduce the network scores on the evaluation inputs. Squared errors are pooled across inputs and actions, and each network action is centered separately:
\setcounter{equation}{1}
\begin{equation}
R^2=1-\frac{\sum_{j=1}^{N}\sum_{a\in\mathcal{A}}(S(x_j,a)-\widehat{Q}(x_j,a))^2}{\sum_{j=1}^{N}\sum_{a\in\mathcal{A}}(S(x_j,a)-\bar{S}_a)^2}\label{eq:B2}
\end{equation}
Here, $N$ is the number of evaluation inputs and $\bar{S}_a=\frac{1}{N}\sum_{j=1}^{N}S(x_j,a)$. The implementation adds $10^{-12}$ to the denominator for numerical stability. Support correlation compares the fitted scores with the controlled support function. Certificate coverage is the fraction satisfying Eq.~\eqref{eq:19}. The real-task score correlations in Appendix E compare fitted scores with network scores.

\subsection{Aggregation and uncertainty}\label{b.4-aggregation-and-uncertainty}

Five training seeds, 42--46, are the independent units. The margin comparison uses 96 test inputs per seed, sampled before scoring. Representation and checkpoint tables report means \ensuremath{\pm} sample standard deviations (SD). Bracketed intervals are Student-$t$ 95\% confidence intervals (CI) over five seed summaries; paired comparisons use the five within-seed differences. Noise replicates are averaged within seeds before inference.

The code package stores the experiment scripts and their comma-separated values (CSV) and JavaScript Object Notation (JSON) summaries. Its recorded environment uses Python 3.13.9, NumPy 2.3.5, and scikit-learn 1.7.2. The appendix reports the controls that determine the interpretation of each result; implementation parameters and individual run outputs remain in that package.

\section{Learned Representations and Case Decomposition}\label{c-learned-representations-and-case-decomposition}
\setcounter{table}{0}
\setcounter{figure}{0}

\subsection{Representation ablation}\label{c.1-representation-ablation}

The ablation tests which representation preserves the information required to recover the case structure of fitted decisions. The training cases and action returns remain fixed within each seed. We compare standardized raw inputs, a random ReLU feature map of width 16, an untrained 32,16 ReLU network, and the final hidden layer of the trained 32,16 network. All four receive the same ridge return-model fit.

Random features use Gaussian weights scaled by the square root of the input dimension and Gaussian biases with standard deviation 0.1. The untrained network uses Xavier-uniform weights and zero biases. Reconstruction and certificate coverage for every representation are evaluated against the same trained network scores. This common reference makes their fidelity values comparable.

\begin{table}[htbp]
\caption{Reconstruction and decision consistency across representations. Mean $\pm$ SD over five seeds. The trained representation achieves near-exact network-score reconstruction ($R^2=0.9998\pm0.0001$), high support recovery ($0.996\pm0.003$), and the highest coverage ($0.943\pm0.026$), while random features recover part of the support structure with lower coverage ($0.404\pm0.110$). The unrounded mean support correlation for the trained representation is 0.9962.}\label{tab:C1}
\begin{center}
\begin{tabular}{@{}
  >{\raggedright\arraybackslash}p{(\linewidth - 6\tabcolsep) * \real{0.2636}}
  >{\centering\arraybackslash}p{(\linewidth - 6\tabcolsep) * \real{0.2636}}
  >{\centering\arraybackslash}p{(\linewidth - 6\tabcolsep) * \real{0.2364}}
  >{\centering\arraybackslash}p{(\linewidth - 6\tabcolsep) * \real{0.2364}}@{}}
\toprule
\begin{minipage}[b]{\linewidth}\raggedright
\textbf{Representation}
\end{minipage} & \begin{minipage}[b]{\linewidth}\centering
\textbf{Network R\ensuremath{^2}}
\end{minipage} & \begin{minipage}[b]{\linewidth}\centering
\textbf{Support r}
\end{minipage} & \begin{minipage}[b]{\linewidth}\centering
\textbf{Coverage}
\end{minipage} \\
\midrule
Raw covariates & \ensuremath{-}1.0648 \ensuremath{\pm} 0.3527 & 0.393 \ensuremath{\pm} 0.053 & 0.030 \ensuremath{\pm} 0.021 \\
Random features & 0.7663 \ensuremath{\pm} 0.1187 & 0.869 \ensuremath{\pm} 0.062 & 0.404 \ensuremath{\pm} 0.110 \\
Untrained network & \ensuremath{-}0.6710 \ensuremath{\pm} 0.2541 & 0.059 \ensuremath{\pm} 0.154 & 0.109 \ensuremath{\pm} 0.032 \\
Trained network & 0.9998 \ensuremath{\pm} 0.0001 & 0.996 \ensuremath{\pm} 0.003 & 0.943 \ensuremath{\pm} 0.026 \\
\bottomrule
\end{tabular}
\end{center}
\end{table}

Learning improves the correspondence between the fitted scores and the trained network (Table~\ref{tab:C1}). The shared ridge objective admits an algebraic case decomposition for every representation, so the reconstruction contrast concerns the information retained in the features. Random features capture part of the smooth support function, but the learned representation more closely preserves the score organization needed for decision consistency.

\begin{table}[htbp]
\caption{Case-action consistency and coefficient sign across representations. Mean $\pm$ SD over five seeds. The untrained network attains Top-30 action consistency of 0.885 despite support correlation of 0.059 in Table~\ref{tab:C1}. No evaluated query has a fully nonnegative coefficient vector, including queries evaluated with the trained representation.}\label{tab:C2}
\begin{center}
\begin{tabular}{@{}
  >{\raggedright\arraybackslash}p{(\linewidth - 6\tabcolsep) * \real{0.2636}}
  >{\centering\arraybackslash}p{(\linewidth - 6\tabcolsep) * \real{0.2545}}
  >{\centering\arraybackslash}p{(\linewidth - 6\tabcolsep) * \real{0.2364}}
  >{\centering\arraybackslash}p{(\linewidth - 6\tabcolsep) * \real{0.2455}}@{}}
\toprule
\begin{minipage}[b]{\linewidth}\raggedright
\textbf{Representation}
\end{minipage} & \begin{minipage}[b]{\linewidth}\centering
\textbf{Top-30 action\\
consistency}\strut
\end{minipage} & \begin{minipage}[b]{\linewidth}\centering
\textbf{Negative mass}
\end{minipage} & \begin{minipage}[b]{\linewidth}\centering
\textbf{Fully nonnegative\\
queries}\strut
\end{minipage} \\
\midrule
Raw covariates & 0.611 \ensuremath{\pm} 0.179 & 0.506 \ensuremath{\pm} 0.011 & 0.000 \ensuremath{\pm} 0.000 \\
Random features & 0.927 \ensuremath{\pm} 0.015 & 0.713 \ensuremath{\pm} 0.086 & 0.000 \ensuremath{\pm} 0.000 \\
Untrained network & 0.885 \ensuremath{\pm} 0.047 & 0.513 \ensuremath{\pm} 0.018 & 0.000 \ensuremath{\pm} 0.000 \\
Trained network & 0.945 \ensuremath{\pm} 0.032 & 0.746 \ensuremath{\pm} 0.034 & 0.000 \ensuremath{\pm} 0.000 \\
\bottomrule
\end{tabular}
\end{center}
\end{table}

\begin{figure}[htbp]
\centering
\begin{tabular}{@{}c@{\hspace{8pt}}c@{\hspace{8pt}}c@{}}
\includegraphics[scale=0.49556599]{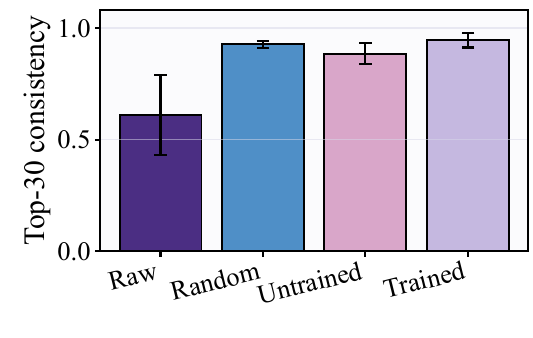} & \includegraphics[scale=0.49556599]{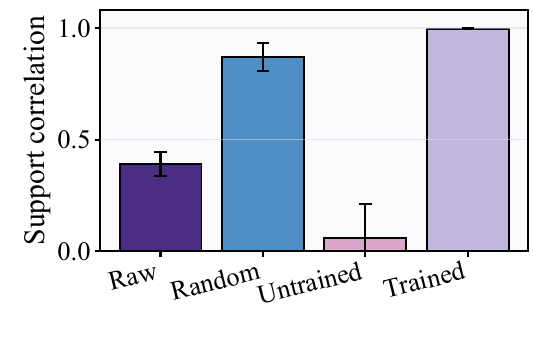} & \includegraphics[scale=0.49556599]{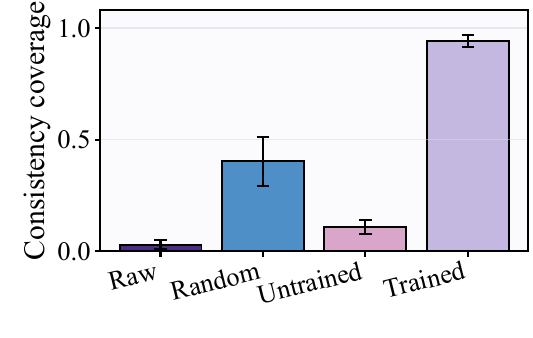}\\
\parbox[t]{126.66667bp}{\centering (a) Action consistency} & \parbox[t]{126.66667bp}{\centering (b) Support correlation} & \parbox[t]{126.66667bp}{\centering (c) Consistency coverage}
\end{tabular}
\caption{Representation ablation on the controlled CBDT task. Panels report (a) weighted Top-30 action consistency, (b) support correlation, and (c) decision-consistency coverage. The untrained network retains high action consistency but weak support correlation; the trained representation has the highest means across all three measures. Error bars show SD across five seeds. Tables~\ref{tab:C1} and \ref{tab:C2} give the numerical results.}\label{fig:C1}
\end{figure}

Retrieving cases with matching preferred actions does not require accurate recovery of their aggregate support. This distinction explains why case-action consistency can remain high for a representation with weak support reconstruction (Table~\ref{tab:C2}; Figure~\ref{fig:C1}). The coefficient signs impose a separate condition: the recovered structure in this setting is a signed case decomposition, even when its decisions and support scores closely match the reference.

\subsection{Changes across training checkpoints}\label{c.2-changes-across-training-checkpoints}

At each checkpoint we recompute the representation matrix and fit fresh return models. The overlap of the largest-coefficient case sets measures how the input-to-case mapping develops relative to the final fit. Comparing this overlap with score reconstruction tests whether faithful predictions and a stable case map emerge at the same stage of learning (Table~\ref{tab:C3}).

\begin{table}[htbp]
\caption{Training-checkpoint trajectory from epoch 1 to epoch 2000. Train MSE is a seed mean; other entries are mean $\pm$ SD over five seeds. Reconstruction reaches $R^2=0.981$ at epoch 500 and approximately 1 at epoch 2000. The Top-30 absolute-coefficient set has mean Jaccard overlap of 0.539 and 0.708 with the final set at epochs 500 and 1000, respectively. Its final overlap of 1 is a self-comparison.}\label{tab:C3}
\begin{center}
\begin{tabular}{@{}
  >{\raggedright\arraybackslash}p{(\linewidth - 8\tabcolsep) * \real{0.0909}}
  >{\centering\arraybackslash}p{(\linewidth - 8\tabcolsep) * \real{0.1236}}
  >{\centering\arraybackslash}p{(\linewidth - 8\tabcolsep) * \real{0.3000}}
  >{\centering\arraybackslash}p{(\linewidth - 8\tabcolsep) * \real{0.2273}}
  >{\centering\arraybackslash}p{(\linewidth - 8\tabcolsep) * \real{0.2582}}@{}}
\toprule
\begin{minipage}[b]{\linewidth}\raggedright
\textbf{Epoch}
\end{minipage} & \begin{minipage}[b]{\linewidth}\centering
\textbf{Train MSE}
\end{minipage} & \begin{minipage}[b]{\linewidth}\centering
\textbf{Network R\ensuremath{^2}}
\end{minipage} & \begin{minipage}[b]{\linewidth}\centering
\textbf{Action\\
agreement}\strut
\end{minipage} & \begin{minipage}[b]{\linewidth}\centering
\textbf{Top-30 coefficient\\
Jaccard}\strut
\end{minipage} \\
\midrule
1 & 184.748 & \ensuremath{-}2668.125 \ensuremath{\pm} 878.130 & 0.109 \ensuremath{\pm} 0.070 & 0.427 \ensuremath{\pm} 0.033 \\
25 & 169.058 & \ensuremath{-}2368.707 \ensuremath{\pm} 3729.723 & 0.528 \ensuremath{\pm} 0.246 & 0.389 \ensuremath{\pm} 0.066 \\
100 & 84.705 & \ensuremath{-}7.864 \ensuremath{\pm} 8.546 & 0.780 \ensuremath{\pm} 0.070 & 0.434 \ensuremath{\pm} 0.013 \\
500 & 2.748 & 0.981 \ensuremath{\pm} 0.009 & 0.935 \ensuremath{\pm} 0.024 & 0.539 \ensuremath{\pm} 0.023 \\
1000 & 0.335 & 0.999 \ensuremath{\pm} 0.000 & 0.985 \ensuremath{\pm} 0.015 & 0.708 \ensuremath{\pm} 0.043 \\
2000 & 0.130 & 1.000 \ensuremath{\pm} 0.000 & 0.992 \ensuremath{\pm} 0.004 & 1.000 \ensuremath{\pm} 0.000 \\
\bottomrule
\end{tabular}
\end{center}
\end{table}

\begin{figure}[htbp]
\centering
\begin{tabular}{@{}c@{\hspace{8pt}}c@{}}
\includegraphics[scale=0.61941252]{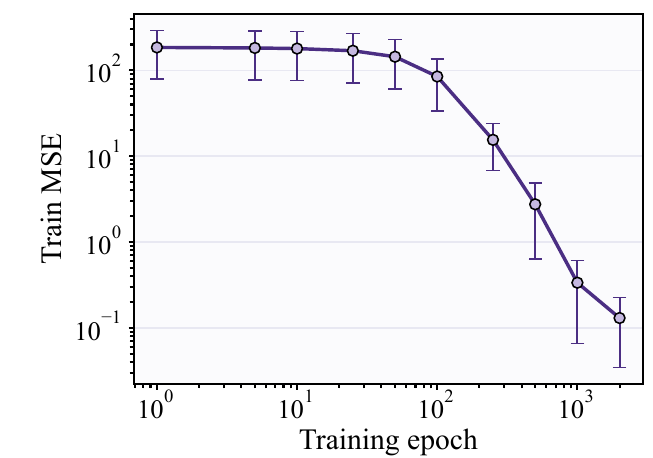} & \includegraphics[scale=0.61941252]{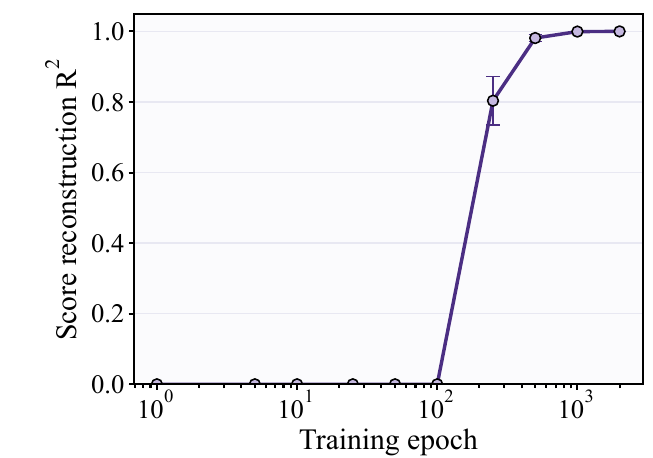}\\
\parbox[t]{194.00000bp}{\centering (a) Training loss} & \parbox[t]{194.00000bp}{\centering (b) Score reconstruction}
\end{tabular}
\par\medskip
\begin{tabular}{@{}c@{\hspace{8pt}}c@{}}
\includegraphics[scale=0.61941252]{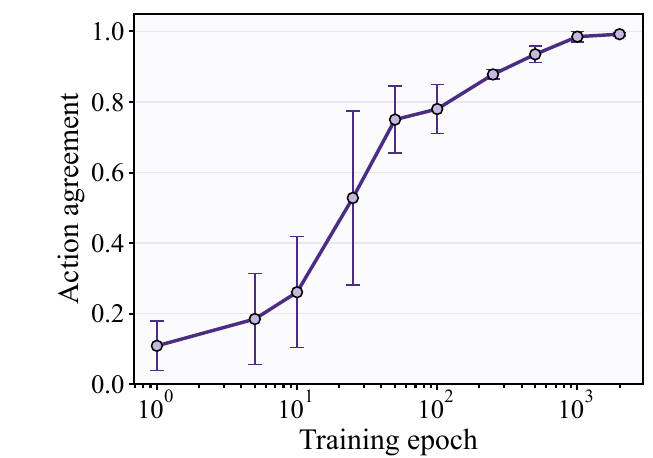} & \includegraphics[scale=0.61941252]{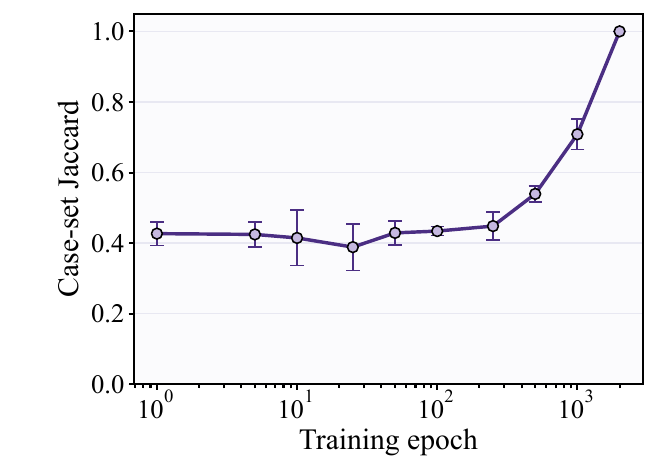}\\
\parbox[t]{194.00000bp}{\centering (c) Action agreement} & \parbox[t]{194.00000bp}{\centering (d) Case-set Jaccard}
\end{tabular}
\caption{Case maps across training checkpoints. (a) Training loss, (b) network-output reconstruction, (c) fitted--network action agreement, and (d) overlap with the final Top-30 absolute-coefficient set. Reconstruction and action agreement improve while the leading case set continues to change. Negative $R^2$ values are displayed at zero in panel (b); Table~\ref{tab:C3} reports their original values. Error bars show SD across five seeds; panel (a) plots the seed-mean MSE.}\label{fig:C2}
\end{figure}

Early reconstruction errors are large relative to the network\textquotesingle s small score variance. As training progresses, accurate score reconstruction precedes stabilization of the cases receiving the largest coefficients (Figure~\ref{fig:C2}). A decision score aggregates many contributions, so similar aggregate scores can be supported by different leading case sets. The trajectory therefore separates fidelity of the recovered scores from stability of the recovered input-to-case mapping.

The checkpoint run retains weights between successive training calls and restarts optimizer state. It is separate from the final-model ablation.

\section{Conditions for Case-Based Interpretation}\label{d-conditions-for-case-based-interpretation}
\setcounter{table}{0}
\setcounter{figure}{0}

\subsection{Comparison with an explicit CBDT Reference}\label{d.1-comparison-with-an-explicit-cbdt-reference}

Figure~\ref{fig:3} compares the neural network with a separate nonnegative CBDT rule on the bidding task. This comparison concerns $U$ and $S$, whereas the fitted-model comparison in Appendix E concerns $\widehat{Q}_\lambda$ and $S$. Distinguishing these pairs matters because agreement with an explicit CBDT rule does not identify the case structure induced by the network\textquotesingle s representation.

The low Aggressive correlation shows that frequent agreement on the winning action can coexist with weak agreement on one action\textquotesingle s score. It does not establish the signs of the representation-induced coefficients. We examine those coefficients directly to evaluate the condition in Section 4.

\subsection{Negative coefficients and the scale of clipping}\label{d.2-negative-coefficients-and-the-scale-of-clipping}

To examine whether the recovered case structure satisfies the case-based interpretation condition, we evaluate 50 randomly sampled test inputs per seed using the wider bidding network in Table~\ref{tab:B1}. For each input, $\nu_{\lambda}(x)=\sum_{i=1}^{n}\max\{-\alpha_{\lambda,i}(x),0\}$ is its negative mass. The negative-coefficient rate is the proportion of negative entries, averaged over queries. Both quantities are summarized in Table~\ref{tab:D1}.

\begin{table}[htbp]
\caption{Coefficient signs on 50 bidding queries per seed. Mean $\pm$ SD across five seeds. All 250 inputs have at least one negative coefficient. The action-averaged clipping bound $E_{\lambda}^{\mathrm{avg}}$ defined in the text has pooled mean 200.4 and median 184.1; its median ratios to the fitted score range and winning margin are 39.0 and 54.0, respectively.}\label{tab:D1}
\begin{center}
\begin{tabular}{@{}
  >{\raggedright\arraybackslash}p{(\linewidth - 2\tabcolsep) * \real{0.6909}}
  >{\centering\arraybackslash}p{(\linewidth - 2\tabcolsep) * \real{0.3091}}@{}}
\toprule
\begin{minipage}[b]{\linewidth}\raggedright
Quantity
\end{minipage} & \begin{minipage}[b]{\linewidth}\centering
\textbf{Value}
\end{minipage} \\
\midrule
Negative-coefficient rate & 0.440 \ensuremath{\pm} 0.007 \\
Mean negative mass & 0.877 \ensuremath{\pm} 0.047 \\
Mean within-seed median negative mass & 0.840 \ensuremath{\pm} 0.050 \\
Mean within-seed 95th percentile negative mass & 1.313 \ensuremath{\pm} 0.171 \\
Fraction of fully nonnegative queries & 0.000 \ensuremath{\pm} 0.000 \\
\bottomrule
\end{tabular}
\end{center}
\end{table}

The sign pattern in Table~\ref{tab:D1} prevents a direct nonnegative-relevance interpretation under the condition in Section 4. The fitted scores still admit the signed decomposition derived from the ridge objective. Nonnegativity therefore determines the CBDT interpretation of the coefficients rather than the validity of the decomposition.

To compare the clipping bound with the fitted score scale, set $R_a=\max_i|r_i(a)|$ and $E_{\lambda}^{\mathrm{avg}}(x)=\nu_{\lambda}(x)\sum_{a\in\mathcal{A}}R_a/|\mathcal{A}|$. This is the action-averaged upper bound in Eq.~\eqref{eq:A10}, evaluated with the ridge coefficients. Its scale relative to the fitted scores leaves the actual clipping error unresolved. The individual decision below therefore retains the signed coefficients.

\subsection{Supporting and opposing cases for one decision}\label{d.3-supporting-and-opposing-cases-for-one-decision}

The following example illustrates how the recovered case structure traces an individual decision back to supporting and opposing cases. For the query in Table~\ref{tab:D2}, let $a_0,a_1,a_2$ denote the low, medium, and high bids, respectively. In this order, the network and fitted score vectors are
\setcounter{equation}{0}
\begin{equation}
S(x,\cdot)=[13.908,\ 7.353,\ -2.357],\label{eq:D1}
\end{equation}
\setcounter{equation}{1}
\begin{equation}
\widehat{Q}_\lambda(x,\cdot)=[14.062,\ 6.068,\ -2.596].\label{eq:D2}
\end{equation}
The fitted winning margin exceeds twice the maximum score discrepancy, so Eq.~\eqref{eq:19} certifies the shared decision. Table~\ref{tab:D2} resolves that margin into supporting and opposing contributions for $a_0$ against the runner-up $a_1$.

\begin{table}[htbp]
\caption{Three supporting and three opposing cases for $a_0$ against runner-up $a_1$, using seed 46 and test input 2325. The unrounded fitted margin is 7.9938 and the maximum score discrepancy is 1.2852. All six displayed coefficients are positive; return contrasts distinguish support from opposition. For case 1221, $0.005749\times(-70.152)\approx-0.4033$. The displayed contributions sum to $-0.7045$; the remaining cases contribute 8.6983, completing the fitted margin.}\label{tab:D2}
\begin{center}
\begin{tabular}{@{}
  >{\raggedright\arraybackslash}p{(\linewidth - 8\tabcolsep) * \real{0.1545}}
  >{\centering\arraybackslash}p{(\linewidth - 8\tabcolsep) * \real{0.1091}}
  >{\centering\arraybackslash}p{(\linewidth - 8\tabcolsep) * \real{0.2455}}
  >{\centering\arraybackslash}p{(\linewidth - 8\tabcolsep) * \real{0.2545}}
  >{\centering\arraybackslash}p{(\linewidth - 8\tabcolsep) * \real{0.2364}}@{}}
\toprule
\begin{minipage}[b]{\linewidth}\raggedright
\textbf{Role}
\end{minipage} & \begin{minipage}[b]{\linewidth}\centering
\textbf{Case}
\end{minipage} & \begin{minipage}[b]{\linewidth}\centering
\textbf{Coefficient}
\end{minipage} & \begin{minipage}[b]{\linewidth}\centering
\textbf{Return contrast}
\end{minipage} & \begin{minipage}[b]{\linewidth}\centering
\textbf{Contribution}
\end{minipage} \\
\midrule
Support & 1773 & +0.005908 & +19.894 & +0.1175 \\
Support & 1452 & +0.005683 & +19.843 & +0.1128 \\
Support & 1312 & +0.005898 & +19.056 & +0.1124 \\
Oppose & 1221 & +0.005749 & -70.152 & -0.4033 \\
Oppose & 3890 & +0.005544 & -63.990 & -0.3547 \\
Oppose & 8396 & +0.005302 & -54.534 & -0.2892 \\
\bottomrule
\end{tabular}
\end{center}
\end{table}

A case supports the selected action through the product of its coefficient and return contrast. Positive coefficients can therefore contribute on either side of the comparison when cases favor different actions (Table~\ref{tab:D2}). Reading the coefficient sign alone would lose this action-dependent meaning.

The balance of the displayed cases also differs from the balance of the full collection. A short list exposes strong contributions in each direction, but the decision follows from their sum over all training cases. The margin accounting in Table~\ref{tab:D2} links that local explanation to the certified choice.

The example was selected from certified inputs with positive-coefficient cases on both sides of the action comparison. The selection requires at least two of the three displayed supporting cases to have positive coefficients and positive return contrasts, and at least one of the three opposing cases to have a positive coefficient and a negative return contrast. This pattern occurs in 8,314 of 15,399 certified inputs across the five wider-network runs. These counts belong to the case-record experiment, whose architecture differs from the consistency experiment in Appendix E.1.

\section{Decision Consistency and Return Error}\label{e-decision-consistency-and-return-error}
\setcounter{table}{0}
\setcounter{figure}{0}

\subsection{Coverage of the sufficient condition}\label{e.1-coverage-of-the-sufficient-condition}

The decision-consistency experiment evaluates Eq.~\eqref{eq:19} on all held-out inputs of each task (Table~\ref{tab:E1}). The certificate compares the winning margin with twice the score error. Its value is therefore input-specific: an error small enough to preserve one decision may change another when competing actions have nearly equal scores.

\begin{table}[htbp]
\caption{Decision consistency across tasks. Means and 95\% seed-level intervals over five seeds. Every certified input has matching fitted and network actions in every task and seed. Coverage ranges from 0.799 to 0.955, while agreement among uncertified inputs ranges from 0.787 to 0.826. The seed-mean ratio of uncertified to overall disagreement is 22.67 for PJM, 12.84 for Adult, and 5.43 for Default.}\label{tab:E1}
\begin{center}
\begin{tabular}{@{}
  >{\raggedright\arraybackslash}p{(\linewidth - 6\tabcolsep) * \real{0.1182}}
  >{\centering\arraybackslash}p{(\linewidth - 6\tabcolsep) * \real{0.2946}}
  >{\centering\arraybackslash}p{(\linewidth - 6\tabcolsep) * \real{0.2946}}
  >{\centering\arraybackslash}p{(\linewidth - 6\tabcolsep) * \real{0.2927}}@{}}
\toprule
\begin{minipage}[b]{\linewidth}\raggedright
\textbf{Task}
\end{minipage} & \begin{minipage}[b]{\linewidth}\centering
\textbf{Overall agreement}
\end{minipage} & \begin{minipage}[b]{\linewidth}\centering
\textbf{Coverage}
\end{minipage} & \begin{minipage}[b]{\linewidth}\centering
\textbf{Uncertified\\
agreement}\strut
\end{minipage} \\
\midrule
PJM & 0.992 {[}0.987, 0.997{]} & 0.955 {[}0.945, 0.965{]} & 0.826 {[}0.753, 0.899{]} \\
Adult & 0.983 {[}0.976, 0.990{]} & 0.919 {[}0.896, 0.942{]} & 0.791 {[}0.757, 0.825{]} \\
Default & 0.953 {[}0.914, 0.991{]} & 0.799 {[}0.721, 0.877{]} & 0.787 {[}0.659, 0.915{]} \\
\bottomrule
\end{tabular}
\end{center}
\end{table}

Agreement among uncertified inputs shows that Eq.~\eqref{eq:19} selects a sufficient subset of matching decisions rather than characterizing every matching decision. The concentration of disagreements outside that subset follows from this partition.

Let $A_{\mathrm{all}}$ and $A_{\mathrm{unc}}$ be agreement rates over all queries and uncertified queries, respectively. The reported disagreement ratio is $(1-A_{\mathrm{unc}})/(1-A_{\mathrm{all}})$ within each seed. Since no certified query disagrees, this ratio is the reciprocal of the uncertified fraction whenever disagreements occur. It describes how the sufficient condition partitions these evaluations.

\subsection{Action-wise score reconstruction}\label{e.2-action-wise-score-reconstruction}

Action-wise reconstruction localizes differences that an aggregate agreement rate can hide (Table~\ref{tab:E2}). In particular, a model can preserve many winning actions while approximating one candidate action less closely. Inspecting the full score vector therefore helps explain where decision consistency may become harder to certify.

\begin{table}[htbp]
\caption{Fitted--network score correlation by action. Means and 95\% seed-level intervals over five seeds. Adult\textquotesingle s mean correlations range from 0.996 to 0.999. Default\textquotesingle s Standard action has a lower mean correlation of 0.956, whereas Reduce and Increase are both approximately 0.998.}\label{tab:E2}
\begin{center}
\begin{tabular}{@{}
  >{\raggedright\arraybackslash}p{(\linewidth - 4\tabcolsep) * \real{0.1818}}
  >{\centering\arraybackslash}p{(\linewidth - 4\tabcolsep) * \real{0.2727}}
  >{\centering\arraybackslash}p{(\linewidth - 4\tabcolsep) * \real{0.5455}}@{}}
\toprule
\begin{minipage}[b]{\linewidth}\raggedright
\textbf{Task}
\end{minipage} & \begin{minipage}[b]{\linewidth}\centering
\textbf{Action}
\end{minipage} & \begin{minipage}[b]{\linewidth}\centering
\textbf{Pearson r {[}95\% CI{]}}
\end{minipage} \\
\midrule
Adult & Conservative & 0.997 {[}0.995, 0.999{]} \\
Adult & Standard & 0.996 {[}0.995, 0.997{]} \\
Adult & Expanded & 0.999 {[}0.998, 1.000{]} \\
Default & Reduce & 0.998 {[}0.997, 0.999{]} \\
Default & Standard & 0.956 {[}0.916, 0.996{]} \\
Default & Increase & 0.998 {[}0.997, 0.999{]} \\
\bottomrule
\end{tabular}
\end{center}
\end{table}

The weaker Standard-action correspondence accompanies the smaller coverage on Default. Correlation summarizes variation across inputs; the certificate uses the maximum absolute score discrepancy at each input. Thus the two measures describe different aspects of fidelity. Figure~\ref{fig:4} in the main text shows their relationship across the tasks.

\subsection{Sensitivity to errors in training returns}\label{e.3-sensitivity-to-errors-in-training-returns}

Estimated returns can change a fitted decomposition even when the representation is held fixed. To isolate this effect on the controlled task, we perturb the training returns by
\setcounter{equation}{0}
\begin{gather}
\tilde{r}_i(a)=r_i(a)+\sigma_r\sigma_a\eta_{i,a}\label{eq:E1}\\
\sigma_a=\sqrt{\frac{1}{n-1}\sum_{i=1}^{n}(r_i(a)-\bar{r}_a)^2}\notag\\
\bar{r}_a=\frac{1}{n}\sum_{i=1}^{n}r_i(a)\notag
\end{gather}

Here, $\tilde{r}_i(a)$ denotes the perturbed return, $\sigma_a$ is the sample standard deviation of the unperturbed training returns for action $a$, and $\sigma_r\in\{0,0.05,0.10,0.20,0.40\}$ controls the noise magnitude. The variables $\eta_{i,a}\sim\mathcal{N}(0,1)$ are independent. Each nonzero level has ten noise replicates per training seed. The network, representation, data split, and test inputs remain fixed. We refit the return models using the perturbed return table and compare them with the original network scores.

The coefficients $\alpha_\lambda(x)$ remain unchanged because they depend on the representation, not on the returns. Case-margin scores change through the perturbed return contrasts. For their Pearson and Jaccard comparisons, we retain each clean model\textquotesingle s selected-versus-runner-up action pair. Replicate-level results are averaged within seeds before forming confidence intervals.

\begin{table}[htbp]
\caption{Sensitivity to errors in training returns on the controlled task. Means and 95\% seed-level intervals over five seeds, with ten noise replicates averaged within each seed at nonzero levels. At $\sigma_r=0.20$, action agreement is 0.964 and contribution correlation is 0.953, while Top-30 Jaccard is 0.687. At $\sigma_r=0.40$, Jaccard decreases to 0.505 and certificate coverage to 0.826. Contribution comparisons retain the clean model\textquotesingle s action pair.}\label{tab:E3}
\begin{center}
\begin{tabular}{@{}lcccc@{}}
\toprule
\textbf{Noise} & \shortstack{\textbf{Action}\\\textbf{agreement}} &
\textbf{Coverage} & \shortstack{\textbf{Contribution}\\\textbf{Pearson}} &
\shortstack{\textbf{Top-30}\\\textbf{Jaccard}} \\
\midrule
0.00 & \shortstack{0.988\\{}[0.977, 0.999]} & \shortstack{0.943\\{}[0.910, 0.976]} & \shortstack{1.000\\{}[1.000, 1.000]} & \shortstack{1.000\\{}[1.000, 1.000]} \\
0.05 & \shortstack{0.984\\{}[0.974, 0.994]} & \shortstack{0.934\\{}[0.906, 0.962]} & \shortstack{0.997\\{}[0.995, 0.998]} & \shortstack{0.901\\{}[0.869, 0.933]} \\
0.10 & \shortstack{0.977\\{}[0.969, 0.985]} & \shortstack{0.918\\{}[0.899, 0.937]} & \shortstack{0.987\\{}[0.981, 0.993]} & \shortstack{0.819\\{}[0.768, 0.871]} \\
0.20 & \shortstack{0.964\\{}[0.953, 0.976]} & \shortstack{0.882\\{}[0.862, 0.901]} & \shortstack{0.953\\{}[0.932, 0.974]} & \shortstack{0.687\\{}[0.620, 0.754]} \\
0.40 & \shortstack{0.945\\{}[0.936, 0.953]} & \shortstack{0.826\\{}[0.806, 0.847]} & \shortstack{0.855\\{}[0.803, 0.908]} & \shortstack{0.505\\{}[0.435, 0.575]} \\
\bottomrule
\end{tabular}
\end{center}
\end{table}

\begin{figure}[htbp]
\centering
\begin{minipage}[b]{0.68\linewidth}\centering
\includegraphics[width=\linewidth]{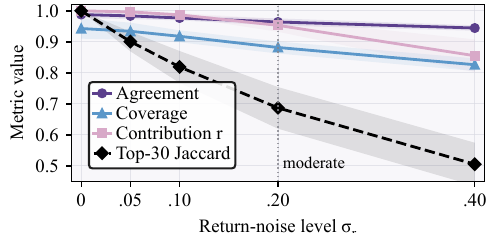}
\end{minipage}
\caption{Sensitivity to training-return noise with the network and representation held fixed. Action agreement and contribution correlation decline more slowly than Top-30 case-set overlap. Certificate coverage also decreases as noise increases. The noise scale $\sigma_r$ is relative to each action\textquotesingle s training-return SD. Table~\ref{tab:E3} reports means and 95\% seed-level intervals.}\label{fig:E1}
\end{figure}

The fitted decision and overall contribution pattern are less sensitive to return perturbations than the identities of the highest-ranked cases (Table~\ref{tab:E3}; Figure~\ref{fig:E1}). Because the representation is fixed, this change arises through the return contrasts rather than the case coefficients. Each perturbed fit retains its algebraic decomposition, while consistency with the original network depends on the score discrepancy produced by the perturbation.

\section{Complementary Comparisons and Computational Cost}\label{f-complementary-comparisons-and-computational-cost}
\setcounter{table}{0}
\setcounter{figure}{0}

\subsection{Comparisons on the controlled margin reference}\label{f.1-comparisons-on-the-controlled-margin-reference}

Table~\ref{tab:2} compares the recovered case contributions from case decomposition with attribution-inspired scores computed over the same fixed representation and fitted linear return models. This comparison examines how the recovered case structure relates to existing training-data attribution formulations under a shared fitted decision model. The comparison uses the action pair and reference in Eq.~\eqref{eq:B1}. These implementations operate on the fitted linear model, so their required state and scoring formulas are specified below.

Let $K_i(x)=\phi(x)^\top\phi(x_i)$ and let $e_i(a,b)$ be the difference of fitted training residuals for actions $a$ and $b$:
\setcounter{equation}{0}
\begin{equation}
e_i(a,b)=[\widehat{Q}_\lambda(x_i,a)-r_i(a)]-[\widehat{Q}_\lambda(x_i,b)-r_i(b)].\label{eq:F1}
\end{equation}
\begin{sloppypar}
The case-decomposition score is $\chi_{\lambda,i}(x;a,b)=\alpha_{\lambda,i}(x)\Delta r_i(a,b)$, with $\lambda=10^{-5}$. The influence-function score is $\phi(x)^{\top}H_d^{-1}\phi(x_i)e_i(a,b)$, where $H_d=(2/n)G+10^{-3}I_H$. The representer-style score is $-K_i(x)e_i(a,b)/(2\lambda_{\mathrm{rep}}n)$ with $\lambda_{\mathrm{rep}}=10^{-3}$. The single-checkpoint gradient score is $0.01K_i(x)e_i(a,b)$, and the inner-product score is $K_i(x)e_i(a,b)$.
\end{sloppypar}

These fixed-representation specializations are inspired by Influence Functions, representer points, and TracIn \citep{Koh2017,Yeh2018,Pruthi2020}, using the shared ridge residuals and one final checkpoint. This comparison is intended to characterize the relationship between recovered case contributions and attribution-inspired scores, rather than to position case decomposition as a general-purpose data attribution method. The three inner-product scores have identical absolute rankings; the negative representer scale reverses their signs.

\begin{table}[htbp]
\caption{Paired differences between case decomposition and fixed-representation comparison scores on the controlled margin reference. Differences are case decomposition minus the comparison score. All eight 95\% intervals lie above zero. Two-sided paired $t$-tests use the five within-seed differences; $p$ values are unadjusted for multiple comparisons.}\label{tab:F1}
\begin{center}
\begin{tabular}{@{}
  >{\raggedright\arraybackslash}p{(\linewidth - 6\tabcolsep) * \real{0.1909}}
  >{\centering\arraybackslash}p{(\linewidth - 6\tabcolsep) * \real{0.2818}}
  >{\centering\arraybackslash}p{(\linewidth - 6\tabcolsep) * \real{0.3909}}
  >{\centering\arraybackslash}p{(\linewidth - 6\tabcolsep) * \real{0.1364}}@{}}
\toprule
\begin{minipage}[b]{\linewidth}\raggedright
\textbf{Metric}
\end{minipage} & \begin{minipage}[b]{\linewidth}\centering
\textbf{Comparison score}
\end{minipage} & \begin{minipage}[b]{\linewidth}\centering
\textbf{Difference {[}95\% CI{]}}
\end{minipage} & \begin{minipage}[b]{\linewidth}\centering
\textbf{Paired p}
\end{minipage} \\
\midrule
Pearson & Influence Functions & 0.623 {[}0.495, 0.751{]} & 0.0002 \\
Jaccard & Influence Functions & 0.266 {[}0.178, 0.355{]} & 0.0011 \\
nDCG & Influence Functions & 0.351 {[}0.227, 0.475{]} & 0.0014 \\
Sign agreement & Influence Functions & 0.330 {[}0.266, 0.394{]} & 0.0001 \\
Pearson & Representer-style & 0.574 {[}0.497, 0.650{]} & \textless0.0001 \\
Jaccard & Representer-style & 0.348 {[}0.235, 0.461{]} & 0.0010 \\
nDCG & Representer-style & 0.545 {[}0.351, 0.740{]} & 0.0015 \\
Sign agreement & Representer-style & 0.433 {[}0.375, 0.491{]} & \textless0.0001 \\
\bottomrule
\end{tabular}
\end{center}
\end{table}

The paired comparisons in Table~\ref{tab:F1} support closer alignment between the recovered contributions and the specified relevance-and-return reference. That reference evaluates case-level margin structure, whereas residual-based attribution scores target changes associated with training examples. The comparison therefore concerns their alignment with this reference under the shared fitted model.

\subsection{Action consistency and case-removal refitting}\label{f.2-action-consistency-and-case-removal-refitting}

The unweighted majority action among the same Top-30 margin-ranked cases agrees with the query\textquotesingle s selected action at rates of 0.883 for Influence Functions, 0.865 for case decomposition, and 0.729 for the other two scores. This case-action statistic gives a different ordering from margin-reference alignment.

Removing the ten highest-ranked supporting cases and refitting gives case decomposition a normalized margin reduction of 0.0019 {[}\ensuremath{-}0.0044, 0.0082{]}. Its differences from random removal and representer-style removal are 0.0015 {[}\ensuremath{-}0.0048, 0.0078{]} and \ensuremath{-}0.0150 {[}\ensuremath{-}0.0366, 0.0065{]}. These intervals do not establish stronger removal effects. Refitting changes both the Gram system and return table; the decomposition describes the original fixed fit.

\subsection{Computational cost}\label{f.3-computational-cost}

For a fixed representation, dense Gram preparation costs $O(nH^2+H^3)$. After factorization, forming all query coefficients costs $O(H^2+nH)$, followed by $O(n)$ to form one pairwise contribution vector. Case ranking adds sorting cost.

With 500 training cases and representation width 16, median cached query times are 0.0126 ms for case decomposition and Influence Functions, and approximately 0.0125 ms for the other two scoring rules. These measurements exclude training and representation evaluation, using 500 measured queries after 50 warm-up queries per seed. Including representation evaluation and Top-30 selection gives per-seed median times of 0.1135--0.1602 ms. Both measurements use an Intel Core i5-14500 central processing unit (CPU), Windows 11, and the Intel Math Kernel Library backend.

\end{document}